\documentclass[twoside,11pt]{article}
\usepackage{jair, theapa, rawfonts}
\usepackage[paperwidth=8.5in, paperheight=11in, centering]{geometry}
\usepackage{graphicx}
\usepackage{url}
\usepackage{amsmath}
\usepackage{multirow}
\usepackage{enumerate}
\usepackage{lastpage}
\usepackage{tabularx}

\jairheading{66}{2019}{297-\pageref{LastPage}}{03/2018}{09/2019}
\ShortHeadings{Type-aware Convolutional Neural Networks for Slot Filling}{Adel \& Sch\"{u}tze}
\firstpageno{297}

\begin{document}

\title{Type-aware Convolutional Neural Networks for Slot Filling}

\author{\name Heike Adel \email heike.adel@cis.lmu.de \\
       \name Hinrich Sch\"{u}tze \email inquiries@cislmu.org \\
       \addr Center for Information and Language Processing (CIS)\\
       LMU Munich, Germany 
       }


\maketitle

  \begin{abstract}
The slot filling task aims at extracting answers for
queries about entities from text, such as 
``Who founded Apple''.
In this paper, we focus on the relation classification
component of a slot filling system.
We propose type-aware convolutional neural
networks to benefit from the mutual dependencies between
entity and relation classification. In particular, we
explore different ways of integrating the named entity types
of the relation arguments into a neural network
for relation classification, including a joint training and a
structured prediction approach. To the best of our knowledge,
this is the first study on type-aware neural networks for
slot filling.
The type-aware
models lead to the best results of our slot
filling pipeline. Joint training performs comparable to structured prediction.
To understand the impact of the different
components of the slot filling pipeline,
we perform a recall
analysis, a manual error analysis and several ablation
studies. 
Such analyses are of particular importance to
other slot filling researchers since the official
slot filling evaluations only assess pipeline outputs.
The analyses show that especially coreference
resolution and our convolutional neural networks have a
large positive impact on the final performance of the slot
filling pipeline.
The presented models, the source code of our system as well as 
our coreference resource is publicy available.
\end{abstract}

\section{Introduction}
\label{sec:Introduction}

Knowledge bases provide structured information about
entities and concepts of the world. They are important
resources for artificial intelligence (AI) and 
natural language processing (NLP)
tasks, such as entity disambiguation, question answering or
information retrieval \shortcite{app1,app2,app3}.  Given a
knowledge base, answering a question like ``Who founded
Apple?''  would require only a simple lookup. Similarly,
an automatic assistant or dialogue system could satisfy the
needs of users more easily with the access to a background
knowledge base.  If a user is, for example, looking for
popular sights nearby, an automatic assistant could look up
points of interest and information about them in a
knowledge base.

Popular large-scale knowledge bases, such as Freebase \shortcite{Freebase} or
Wiki\-ped\-ia (\citeauthor{wikiurl}-URL) are often
created in a large collaborative effort.  Despite a lot of
(manual) effort spent on their creation and maintenance,
they are usually incomplete.  Missing facts, however, limit
their applicability in down-stream tasks.  At the same time,
there is a lot of unstructured text data available -- e.g.,
on the internet -- that mentions information missing in
knowledge bases.  Therefore, automatic methods for
extracting structured information from text data to populate
knowledge bases are important.

One specific incarnation of knowledge base population (KBP) is
slot filling \cite{overviewSF2013,overviewSF2014}, a
shared task (\citeauthor{sfurl}-URL) which is 
yearly organized by the Text Analysis Conference
(TAC).
Given a large document collection and a query like
``X founded Apple'', the task is to extract 
``fillers'' for the slot ``X'' from the document
collection.
The extraction of answers to the queries from large
amounts of natural language text involves a variety
of challenges, such as document retrieval,
entity identification, coreference resolution or
cross-document inferences.
To cope with those challenges, most slot filling systems
are pipelines of different NLP components.
One of the most important components 
validates whether a candidate  (e.g., ``Steve
Jobs'') is a correct filler of the slot (e.g., ``X'' in
``X founded Apple'').
We take a \emph{relation classification} approach to
candidate validation in this paper. For example, the relation
classifier validates whether the relation between the noun
phrases ``Steve Jobs'' and ``Apple'' in the sentence ``Steve
Jobs started Apple'' is the
relation ``X founded Y''. 
Traditional methods to slot filling relation classification
rely on (hand-crafted) patterns or linear classifiers
with manually designed features.
Given the variability of language, it is desirable to learn
relation-specific characteristics automatically from data instead.
Therefore, we design convolutional
neural network architectures for the special characteristics
of the slot filling task (e.g., long sentences, many
inverse relations) which learn to recognize
relation-specific n-gram patterns.

\subsection{Contributions}
\label{sec:contributions}
We now describe our contributions in this paper
which are centered around \emph{convolutional neural network} (CNN)
architectures for \emph{relation classification} in the
context of \emph{slot
filling}.

\subsubsection{Architectures and Extensive Experimentation Using
  Convolutional Neural Networks for Slot Filling}  
We were one of the first 
groups to use CNNs for relation classification and
demonstrate their effectiveness for slot filling
\cite{adelNaacl2016,vuNaacl2016}. The system based on this
work (which is described in detail in Section \ref{sec:overview}) 
is state of the art for distantly supervised slot
filling \cite{cis2015}, see Section \ref{sec:soaSF}.  In contrast to
scenarios where carefully labeled gold training sets are
available, relation classifiers in slot filling are trained
on data that is noisy -- due to error propagation through
the pipeline and due to distant supervision
\cite{cis2015,adelNaacl2016}.  We show that CNNs are robust
enough to be successfully applied in this noisy
environment if the generic CNN
architecture is adapted for relation classification and if
hyperparameters are carefully tuned on a per-relation basis 
(see Section \ref{sec:neuralNets}).
We also show that multi-class CNNs perform better than
per-relation binary CNNs in the slot filling pipeline (Section \ref{sec:end-to-end})
probably because imposing a 1-out-of-k constraint models the
data better -- even though there are rare cases where more
than one relation holds true.

\subsubsection{Type-Aware Relation Classification} 
We define relation classification as the problem of
assigning one of several relations to a 5-tuple
$(c_{-1},e_1,c_0,e_2,c_{+1})$ consisting of two entites
$(e_1$, $e_2)$ and preceding, central and following
contexts $(c_{-1}$, $c_0$, $c_{+1})$. 
If entities are represented as they occur in the raw text, the classifier
is likely to overfit to the idiosyncrasies of the entities mentioned
in the training data. 
On the other hand, removing all entity
information from the input is also harmful since entities
provide valuable information for disambiguating relations;
consider  ``Apple launches iPhone'' (to start
selling) vs.\ ``SpaceX launches Falcon 9'' (to send into
orbit). A major focus of this work is \emph{type-aware
  relation classification}, a middle ground between complete
entity information and no entity information: only the
predicted types of the entities are made available. Type information is
arguably the key information needed for disambiguation in
relation classification -- e.g., this is the case for our
``launch'' example); and it prevents overfitting to entity
idiosyncrasies. We design three type-aware architectures: a
simple pipeline of first type classification and then
relation classification (see Section \ref{sec:pipeline}); a joint model (Section \ref{sec:jointTraining}); 
and a structured prediction model that more directly takes into account the
dependencies between entity and relation classes (Section \ref{sec:structuredPrediction}).
In our experiments in Section \ref{sec:end-to-end}, we show
that the structured prediction model outperforms the other models in terms of macro
$F_1$, the best measure of performance for difficult cases
because it gives equal weight to rare and frequent
relations.

\subsubsection{Analysis} 
The TAC KBP organizers only evaluate the final results of
the entire slot filling pipeline.  We perform an extensive
and detailed analysis (Sections \ref{sec:recallAnalysis} and \ref{sec:errorAnalysis}) 
and several ablation studies (Section \ref{sec:ablation}) on
individual modules of the pipeline. Inter alia, we quantify the impact
of entity linking, coreference resolution and type-aware CNNs 
on the overall pipeline performance.
We hope that this  will be of
great benefit to the community because this analysis
-- in contrast to the official TAC KBP evaluation --
 allows researchers to
assess the impact of individual components and which
components are worth investing more research effort in.

\subsubsection{Resources} 
We make our complete slot filling
system, including the source code, publicly available at 
\url{http://cistern.cis.lmu.de/CIS_SlotFilling}.  Since slot
filling poses many NLP challenges, building such
a system is a substantial software development and research effort.
Through publication of the system, we share our experience
with the community and lower the barriers to entry for 
researchers wishing to work on slot filling.

The component of our pipeline that has the longest runtime --
a runtime of several months on the entire TAC source corpus -- is coreference
resolution. Therefore, we publish the output of the
coreference resolver \shortcite<Stanford \textsc{CoreNlp} by>{coreNLP} for the two million documents of the
TAC source corpus at \url{http://cistern.cis.lmu.de/corefresources}. 
It consists of 198 million mentions linked
in 54 million coreference chains. This will make it easier
for other researchers to take advantage of coreference
resolution in their systems.

By making these resources available to the NLP community,
we aim
to promote research
in knowledge base population in general and slot filling in particular.

\subsection{Relation to Our Prior Publications}
In this subsection, we delineate our contributions
in this paper from our prior publications.

In earlier work \cite{cis2015,adelNaacl2016}, we compared binary convolutional
neural networks to traditional models for slot filling
(patterns and support vector machines).  Binary models
facilitate extensions of the slot list with a few more slots
since new models can be trained for the new slots but the
existing models do not need to be retrained. However, the
more slots there are, the more models need to be optimized,
maintained and evaluated.  Therefore, we explore multi-class
convolutional neural networks for slot filling in this
paper. In our experiments, we compare the novel multi-class
models to the previously trained binary models.

The contribution of type-aware models which lead to our
best results on the official slot filling evaluation data,
is entirely novel to this paper. Although entity types are
well-studied features for traditional slot filling models
\shortcite<e.g.,>{stanford2014,cmu2014},
this is the first work to explore end-to-end 
type-aware neural networks for slot filling and
show their positive impact in the pipeline setting.

Our three approaches for type-aware neural networks build on 
models we proposed earlier
\cite{EACLnoiseMitigation,adelEmnlp2017}.
In contrast to these two prior studies, we adapt the 
architectures of the models to the requirements of the
slot filling task, e.g., making them more robust against
unknown rare test entities and against the existence of
inverse relations.
This is also the first work to evaluate the CNNs with structured prediction
in a noisy scenario which is arguably conceptually different
to both clean data with manual annotations and distantly supervised
data used without pipelines. 
For structured prediction, we formalize the task of joint entity and relation
classification as a triple of predictions (similar to a knowledge base triple) which enables
the model to learn which entity and relation classes often
co-occur together. This is a novelty to slot
filling which has been approached mainly with pattern matching or classification so far. 

\subsection{Structure}
The remainder of the paper is organized as follows. First, 
the slot filling task and its challenges are described (Section \ref{sec:task}).
Section \ref{sec:overview} presents our slot filling system
which we used in the official shared task competition in 2015.
In Section \ref{sec:neuralNets}, we describe our
convolutional neural network for slot filling relation classification
and introduce multi-class models as well as models 
for the joint task of entity and relation classification.
Afterwards, we present our experiments and discuss our results in Section \ref{sec:results}. 
Section \ref{sec:analysis}
provides the results of a recall analysis, 
a manual categorization of the errors of our system and
several ablation studies.
Section \ref{sec:relWork} presents related work.
Finally, Section \ref{sec:conclusion}
concludes the paper.

\section{Slot Filling}
In this section, we describe the slot filling task and its challenges
and present the most important aspects of our slot filling system.

\subsection{Task and Challenges}
\label{sec:task}
The TAC KBP slot filling task addresses the challenge of gathering information
about entities (persons, organizations or geo-political entities) 
from a large amount of unstructured text data \cite{overviewSF2013,overviewSF2014}.
The input is a query similar to the following one (fictional query 
with a random query id and document id):
\begin{verbatim}
   <query id="CSSF15_ENG_012abc3456">
    <name>Apple</name>
    <docid>NYT_ENG_20131203.4567</docid>
    <beg>222</beg>
    <end>226</end>
    <enttype>org</enttype>
    <slot0>org:founded_by</slot0>
    <slot1>per:date_of_birth</slot1>
  </query>
\end{verbatim}
This sample query asks for the founders of the company Apple as well as for their dates of birth.
It consists of a unique identifier (\texttt{query id}), the name of an
entity (\texttt{name}), which we will call query entity in the following, and the type of this 
entity (\texttt{enttype}) which can be either person, organization
or geo-political entity. Furthermore, it contains the slots to be filled (\texttt{slot0, slot1}), i.e.,
the questions that should be answered for the query entity,
as well as a starting point in the document collection (\texttt{docid} with
begin (\texttt{beg}) and end offset (\texttt{end})) which points to a mention of the query entity. 
The starting point usually does not provide the answer to the query but it 
can be used to disambiguate different entities with the same name.

\begin{figure}[h]
\centering
 \includegraphics[width=.8\textwidth]{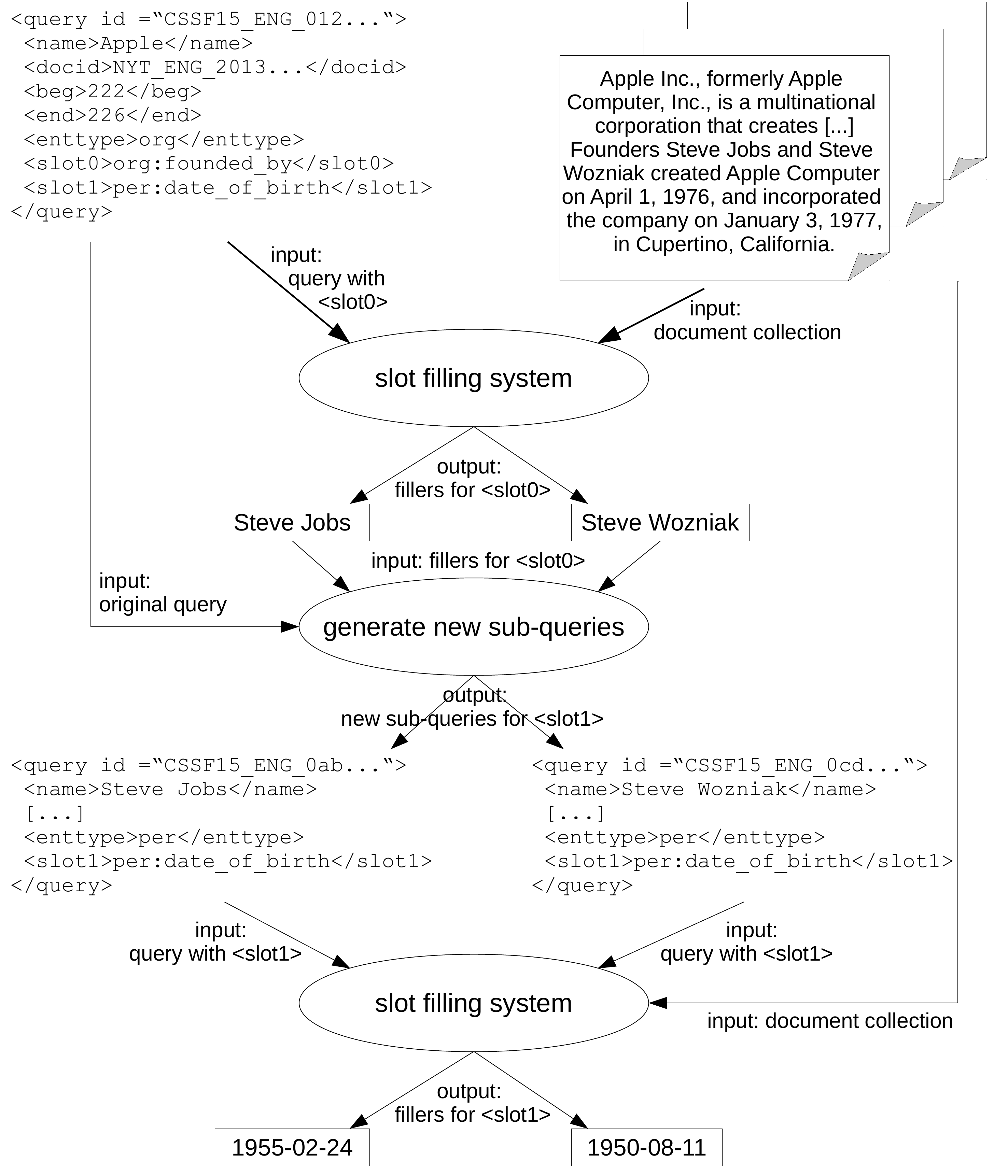}
 \caption{Overview of the slot filling task with a multiple-hop query.}
 \label{fig:task}
\end{figure}

Figure \ref{fig:task} illustrates the slot filling task which is described in the following.
The system has access to a large document collection which needs to be processed
in order to answer the query -- in the example, to find
the founders of Apple and their dates of birth. 
The query can be split into two parts: First, the question posed by
\texttt{slot0} (``Who are the founders of Apple?'') needs to be answered. Based on the results
of the system on this slot, \texttt{slot1} is processed (``What is their date of birth?'').
Since this corresponds to taking a hop in the corresponding knowledge graph
(from Apple to its founders to their date of birth), a query with two
slots is also called multiple-hop query, and the question posed by
\texttt{slot0} is called hop 0 and the question posed by \texttt{slot1} is called hop 1.
The shared task organizers provide a script to generate new sub-queries
for hop 1 given the system results for hop 0 and the original set of queries.
The output of the system should contain the answer for each 
given slot (e.g., ``Steve Jobs'' and ``Steve Wozniak'' for \texttt{org:founded\_by}),
a supporting sentence from the document collection (e.g., 
``Founders Steve Jobs and Steve Wozniak created Apple Computers on April 1, 1976'')
as well as a confidence score.
The slots can be single-valued (for instance, \texttt{per:date\_of\_birth}: a person
has only one date of birth) or list-valued (for instance, \texttt{org:founded\_by}:
a company might have more than one founder).
In total, there are 65 slots, out of which 18 are single-valued and 47 are list-valued.
The answer of the slot filling system is assessed as correct if both
the slot filler and the supporting sentence are correct.
In the official evaluations, human annotators assess system outputs manually.
Based on these assessments, individual results for both hops are reported
as well as an overall result.

Previous work on slot filling showed that this task includes a variety
of NLP challenges \cite{analysisRecall,analysis2012,overviewSF2014}, 
such as alternate names for the same entity, ambiguous names
(i.e., the same name for different 
entities), misspellings, coreference resolution, location inference, 
cross-document inference and relation extraction / classification.
Our slot filling system addresses most of these challenges 
(except for cross-document inference which we only consider
in the context of location inference).

\subsection{The CIS Slot Filling System}
\label{sec:overview}
\begin{figure}[h]
\centering
\includegraphics[width=0.7\textwidth]{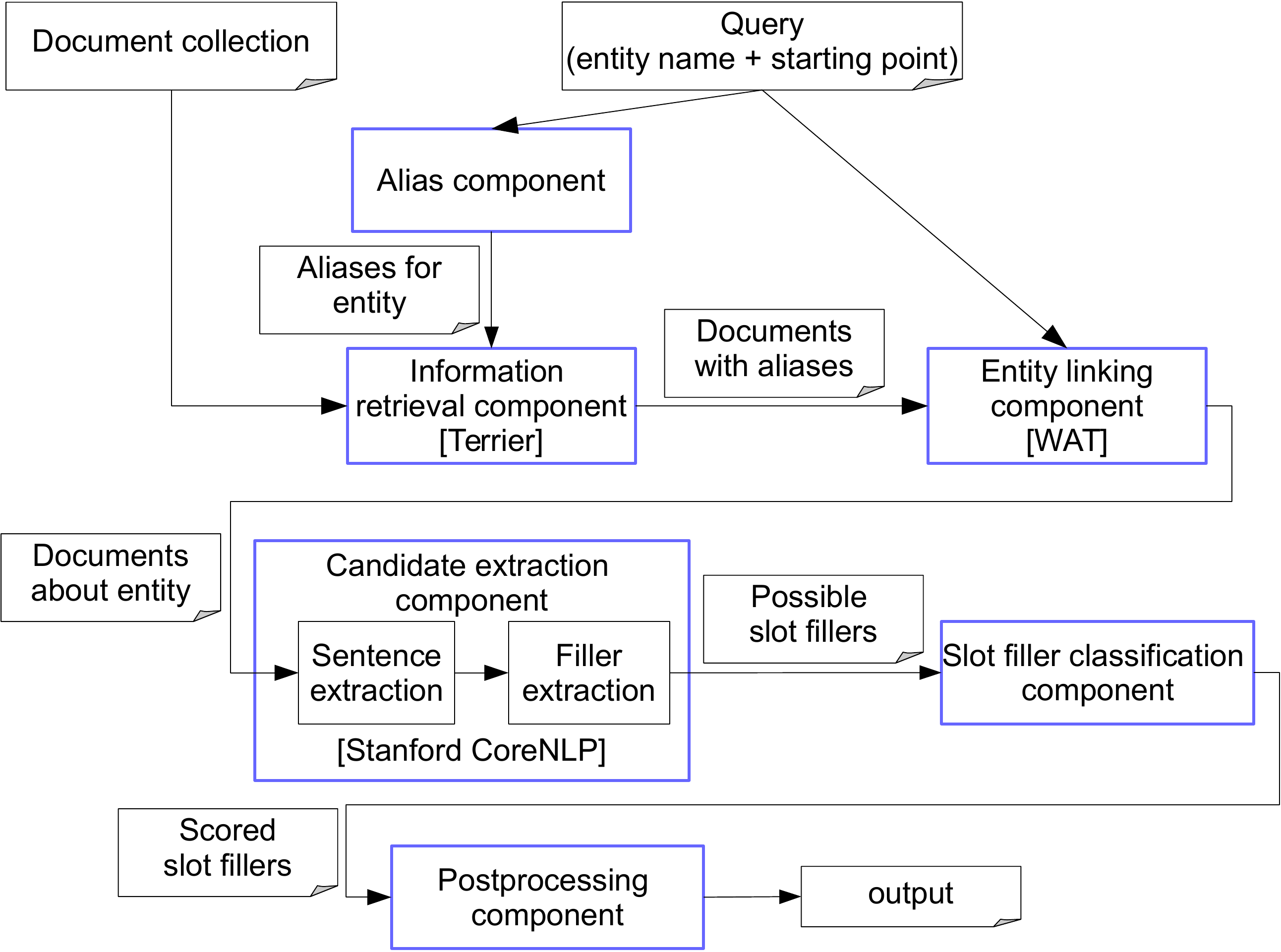}
\caption{System overview: Basic components of the CIS slot filling system.}
\label{fig:systemoverview}
\end{figure}

For filling slots for persons, organizations and geo-political entities, i.e.,
for answering the questions posed by the input queries, a variety of
natural language processing steps need to be performed.
Our system addresses the slot filling task in a modular way. 
This has several advantages, including 
extensibility, component-wise analyzability (see Section \ref{sec:errorAnalysis}) 
and modular development.
In this section, an overview of the different components of our system
is given. They are also depicted in Figure~\ref{fig:systemoverview}.
More details can be found in our shared task system description paper \cite{cis2015}.

\subsubsection{Alias component}
\label{component:alias}
The alias component 
expands the query with possible aliases
for the entity name. 
For this purpose, we employ a preprocessed list of possible aliases based on Wikipedia
redirects which we extracted using \textsc{Jwpl} \cite{JWPL} on a Wikipedia dump from July 2014.
If the query entity is an organization, we also add various 
company-specific suffixes to the list of aliases, such as ``Corp'', ``Co'', ``Inc''.
If the query entity is a person, we include nicknames 
taken from the web (\citeauthor{nicknameurl}-URL)
into the list of aliases.

\subsubsection{Information retrieval component}
\label{component:IR}
Based on the entity name (and aliases),
documents mentioning this name are retrieved to reduce the large
search space. 
For this, we apply the open-source
system \textsc{Terrier} \shortcite{Terrier} with
the following set of queries:
\begin{itemize}
 \item AND combination of the tokens of the entity name as given in the input query
 \item AND combination of the tokens of an alias
 \item OR combination of the tokens of the entity name as given in the input query
\end{itemize}
For geo-political entities, we only use the two AND queries.
In prior experiments, we also investigated phrase queries but found that
they did not work well with spelling variations, resulting in a
considerably lower overall recall of the system. 
Instead, we filter the resulting list of relevant documents
by fuzzy string matching with the name and aliases to skip
documents mentioning both the first and the last name of a
person but not in a phrase.

For each entity, the results of the subqueries are ordered
according to the relevance score assigned by \textsc{Terrier}, concatenated
and limited to the top 300 documents.

\subsubsection{Entity linking component}
\label{component:entityLinking}
For disambiguating entities with the same name,
we apply the entity linking system \textsc{Wat} \cite{WAT}.
It takes a sentence as input and determines the Wikipedia id
of every entity in that sentence.
In order to get the Wikipedia id of the query entity, we 
apply it to the sentence specified by the starting point of the query.
Afterwards, we check for each document
returned by the information retrieval component
whether the mention in the document which matches
the name of the query entity refers to the
same Wikipedia entity as the query, i.e., whether \textsc{Wat} assigns it
the same Wikipedia id as the query entity.
In the case of a mismatch, the document is ignored in the following steps.
After entity linking, we limit the set of documents to the 100 documents
with the highest relevance score according to \textsc{Terrier}.
This number has been determined empirically in prior experiments:
On data from previous slot filling evaluations (2013 and 2014), 
we observed that 100 documents are a good 
trade-off between recall and processing time.

\subsubsection{Candidate extraction component}
\label{component:candidateExtraction}

From the remaining documents,
possible slot fillers (filler candidates) are extracted.

\paragraph{Genre-Specific Document Processing.}
First, the documents are split into sentences
(with Stanford \textsc{CoreNlp}) and cleaned, 
i.a., by removing html tags.
Since the TAC 2015 evaluation corpus consists of two 
genres (news and discussion forums),
our document processing and cleaning steps are genre-dependent. 
Prior analysis showed that 
this is crucial to reduce the noise
in the input to the following pipeline components.

\paragraph{Coreference Resolution.}
Second, fuzzy string matching (based on Levenshtein distance)
and automatic coreference resolution (with Stanford \textsc{CoreNlp})
is performed in order to 
retrieve sentences mentioning the query entity.

Different studies show the importance of coreference
resolution for slot filling \cite{analysis2012,analysisRecall,overviewSF2014}.
While most systems apply coreference resolution only for matching the 
query entity, we also use it for the filler candidates if the named entity
type of the filler is \texttt{PERSON}. 
This improves the recall of the system considerably
 (e.g., consider the slot \texttt{org:students} and 
 the sentence ``He went to University of Munich'').
In Section \ref{sec:ablationCoref}, we show the positive impact of coreference
resolution on the slot filling pipeline results.

Given only raw text data, the runtime of the slot filling system
is mainly determined by coreference resolution. Therefore, we
preprocessed the TAC source corpus (which consists of 
over two million documents)
and stored the coreference information. In total, we have 
extracted about 54M coreference chains with a total number
of about 198M mentions. The processing of all documents
of the source corpus takes a considerable amount of time and
may be infeasible in case of restricted computational resources.
Therefore, we make this resource
publicly available to the community (see Section \ref{sec:contributions}).

\paragraph{Filler Candidate Extraction.}
Given sentences with mentions of the query entity, the 
system extracts possible filler candidates based 
on named entity tags. For named entity tagging, we apply \textsc{CoreNlp}.
For example, for the slot \texttt{per:date\_of\_birth},
the system would only consider dates as filler candidates
while for a slot like \texttt{org:members}, the fillers can be organizations,
locations or persons.
For slots with string fillers, such as \texttt{per:title} or 
\texttt{per:charges}, we have automatically assembled lists of possible filler
values from Freebase \cite{Freebase}. The lists have been 
cleaned manually in order to improve their precision.

\subsubsection{Slot Filler Classification Component}
\label{component:classificationComponent}
The classification component identifies valid fillers for the given slot
based on the textual context of the extracted filler candidates.
This is mainly a relation classification task with the additional challenges
that no designated training data is available and that the
classifier inputs are the results from previous pipeline steps
and can, thus, be noisy (e.g., due to wrong coreference resolution, 
wrong named entity typing or erroneous sentence splitting).
Given our previous results \cite{adelNaacl2016},
we combine traditional models \shortcite<patterns and support vector machines (SVMs), similar to
the ones used by>{roth2013}
with convolutional neural networks (CNNs) in this component.
The weights for combining the model scores are tuned
on previous slot filling evaluation data.
Section \ref{sec:neuralNets} describes the CNNs in more detail.

\subsubsection{Postprocessing component}
\label{component:postprocessing}

As a last step, the filler candidates
are postprocessed and output along with the confidence scores from the classification
component and the contexts they appear in. 
\paragraph{Output Thresholds and Ranking.}
Based on the classification scores, filler candidates are selected
for output or discarded. This decision is done based on slot-specific
thresholds which were tuned on previous evaluation data.
For the second slot (hop 1) of multiple-hop queries, we 
increase the thresholds by 0.1 in order to reduce false positive answers.
(An answer to a hop 1 sub-query is only scored as correct if both the hop 0 answer 
and the hop 1 answer are correct. Thus, errors are propagated from hop 0 to hop 1.)
The selected filler candidates are ranked according to their classification score.
For single-valued slots, only the top filler candidate is output. For list-valued
slots, the top $N$ filler candidates are output. ($N$ is
slot-dependent and has been determined heuristically on previous evaluation 
data in order to increase the precision of the system.)
\paragraph{Location Disambiguation and Inference.}
As we will describe below in Section \ref{sec:neuralNets-remarks}, we do not 
distinguish between cities,
states-or-provinces, and countries in the classification component.
Before outputting the results, however, the extracted locations need to be
disambiguated. The system uses city-, state- and country 
lists from Freebase, Wikipedia (\citeauthor{statelisturl}-URL) 
and an online list of countries (\citeauthor{countrylisturl}-URL)
 to decide to which category a location belongs.
If the system extracted a city or state while the slot given in the query
is a state or country, the system automatically infers the answer for
the desired slot based on city-to-state, city-to-country and state-to-country
mappings extracted from Freebase.

\section{Convolutional Neural Networks for Slot Filling Relation Classification}
\label{sec:neuralNets}

Convolutional neural networks (CNNs) have been applied successfully 
to natural language processing in general
(\shortciteR{cw}; \citeR{kalchbrenner}) and relation 
classification in particular (\shortciteR{zeng}; \citeR{dosSantos,vuNaacl2016}).
We integrate them into a slot filling pipeline. 
This poses two additional challenges to them which prior work usually
does not consider in combination: noisy labels at training time due
to distantly supervised training data; and noisy or wrong inputs
at test time due to error propagations in the slot filling pipeline.
Examples for the latter are wrong sentence boundaries (resulting
in incomplete or very long inputs), wrong coreference resolution or
wrong named entity tags (resulting in incorrect candidate entites for 
relation classification).
Our results show that CNNs are still able to classify the relations and improve
the final performance of the system.

There are three reasons why CNNs are promising models for slot filling
relation classification:
(i) Convolutional filters of length $n$ automatically create features
for every possible $n$-gram in the sentence. Since relation-indicative phrases are
often $n$-grams (examples: ``was born in'' or ``subsidiary of''),
each convolutional filter can learn to recognize a particular $n$-gram and
assign a high score to it. 
(ii) Max pooling, i.e., only considering the highest activations
from each filter application result, helps extracting the most relevant $n$-grams 
independent of their position in the sentence. Thus, the following network layers
can focus on those most relevant parts of the sentence only.
(iii) The representation of input words with word embeddings and
the internal computation of phrase and sentence representations
based on them enables the network to recognize words or phrases
which are similar to the ones seen during training. Thus, if a convolutional
filter has learned to assign high scores to the $n$-gram ``was founded by'' during training,
it can also recognize a phrase like ``was established by'' during testing even if this phrase
did not occur in the training data (assuming that the embeddings of ``founded'' and ``established''
are similar). This is an advantage of neural models compared to pattern
matching or bag-of-word approaches for which this generalization is more difficult.

\subsection{General Remarks}
\label{sec:neuralNets-remarks}
For training and evaluating the convolutional neural networks,
we replace the query entity with the tag \texttt{<name>} and
the candidate filler with the tag \texttt{<filler>}. This prevents
the model from remembering entities from the training data and
helps it to focus on the context words instead.
Furthermore, we only train one model for each
slot and its inverse, for example \texttt{per:parents} and \texttt{per:children}. 
For this, we transform all training examples of \texttt{per:parents} into
training examples for \texttt{per:children} by reversing the \texttt{<name>} and \texttt{<filler>}
tags in the sentences.
This avoids redundant training and helps the model to discriminate between
the two inverse slots.
To extract the probability for 
an inverse slot during test time (e.g., for \texttt{per:parents}),
we again reverse the \texttt{<name>} and \texttt{<filler>}
tags of the sentence and extract the probability for the corresponding slot (e.g., \texttt{per:children}).
We also merge the ``city'', ``country'' and ``state-or-province'' slots 
to one ``location'' slot since we expect their fillers to appear in very similar contexts.

\subsection{Basic Architecture}
This section describes the basic architecture of our convolutional
neural network for slot filling relation classification (without entity types).

\subsubsection{Model Input}
\label{sec:neuralNets-input}
We only use features which are directly available
from the input context, i.e., words and combinations of words, but no 
hand-crafted features, such as part-of-speech tags or dependency paths.
One reason for that is to avoid potential noise in
the inputs due to wrong tags 
or wrong dependency paths. As described before, the slot filling
pipeline introduces different kinds of noise anyway. Therefore,
we aim for limiting additional noise as much as possible.
Another reason is previous results: In earlier work \cite{adelNaacl2016},
we showed that models without dependency paths as features
are able to outperform models using dependency paths (\textsc{Mintz++} and \textsc{MimlRe})
on slot filling relation classification.
Similarly, \citeA{roth2013} achieved the best results in the slot filling
evaluations with only a minimalistic feature set.

We split the input sentence into three parts: left of the candidate relation
arguments (left context), between the arguments (middle context) 
and right of them (right context). These three contexts form the input
to the CNN, together with a flag indicating whether the query entity or the candidate filler
 appears first in the sentence. This flag is important for disambiguating
 inverse relations (see Section \ref{sec:neuralNets-remarks}).

\subsubsection{Convolutional Layer}

The words of the input sentence are represented with
word embeddings pre-trained with
word2vec \shortcite{word2vec} on English Wikipedia.
As described in Section \ref{sec:neuralNets-input},
the input sentence is split into three parts: left, middle and right context.
The network applies the following Equations \ref{eq:cnn1}-\ref{eq:cnn6}
for convolution and 3-max pooling \cite{kalchbrenner} to each of these
three parts individually but with convolutional filters $H$ and bias terms $b$ shared across the 
three contexts.

\begin{align}
 C_{\text{left}} & = \tanh(I_{\text{left}} \ast H + b)  \label{eq:cnn1}\\
 C_{\text{middle}} & = \tanh(I_{\text{middle}} \ast H + b)  \label{eq:cnn2}\\
 C_{\text{right}} & = \tanh(I_{\text{right}} \ast H + b)
 \label{eq:cnn3}
\end{align}
The symbol $\ast$ denotes convolution, $I$ is an input context, 
$H$ is the filter matrix and $b$ the bias term. 
We apply Equations \ref{eq:cnn1}-\ref{eq:cnn3} with multiple filter matrices $H$.
The number of filter matrices $m \in \left\{100, 300, 1000\right\}$ 
is tuned on the development set.
After convolution, 3-max pooling is applied which extracts the three
maximum values of each $C$ (in the same order as they appeared in the 
input sequence), yielding the pooling results $P_{\text{left}}, P_{\text{middle}},
P_{\text{right}}$:
\begin{align}
 P_{\text{left}}[i] & = [C_{\text{left}}[i,t] | R_t(C_{\text{left}}[i,t]) \le 3] \label{eq:cnn4}\\
 P_{\text{middle}}[i] & = [C_{\text{middle}}[i,t] | R_t(C_{\text{middle}}[i,t]) \le 3] \label{eq:cnn5}\\
 P_{\text{right}}[i] & = [C_{\text{right}}[i,t] | R_t(C_{\text{right}}[i,t]) \le 3] \label{eq:cnn6}
\end{align}
with $[i]$ denoting the $i$-th row and $[i,t]$ the cell in the $i$-th row
and $t$-th column of a matrix. The sequence $[p | P(p)]$ contains
all elements $p$ that satisfy predicate $P$, in this case all elements
whose rank $R_t$ along the time axis $t$ is 1, 2 or 3. 
For more details on k-max pooling, see \citeA{kalchbrenner}.

Because of convolution and pooling, 
the network is able to recognize relevant n-grams independent 
of their position in the input sentence.
Afterwards, the results are concatenated to one large vector and extended 
with a flag $v$ indicating whether the entity or the filler candidate appears 
first in the sentence. The final vector is passed to a
multi-layer perceptron with one hidden layer. 

\begin{equation}
 s = \tanh(W_1^\top P_{\text{left}} + W_2^\top P_{\text{middle}} + W_3^\top P_{\text{right}} + W_4^\top v + d) 
 \label{eq:hidden}
\end{equation}
The matrices $W_1$, $W_2$, $W_3$ and $W_4$ are the weights of the hidden layer,
$d$ is its bias term.

\subsubsection{Output Layer}
Finally, a softmax layer is applied to the sentence representation $s$.
In earlier work \cite{cis2015,adelNaacl2016}, we only trained binary models
which output 1 if $s$ expresses
the given slot or 0 if it does not.
In this paper, we explore multi-class models.
In the case of multi-class models, the output vector contains one 
output entry for each slot (except for inverse slots, 
see Section \ref{sec:neuralNets-remarks}).

\begin{figure}[h]
\centering
\includegraphics[width=0.6\textwidth]{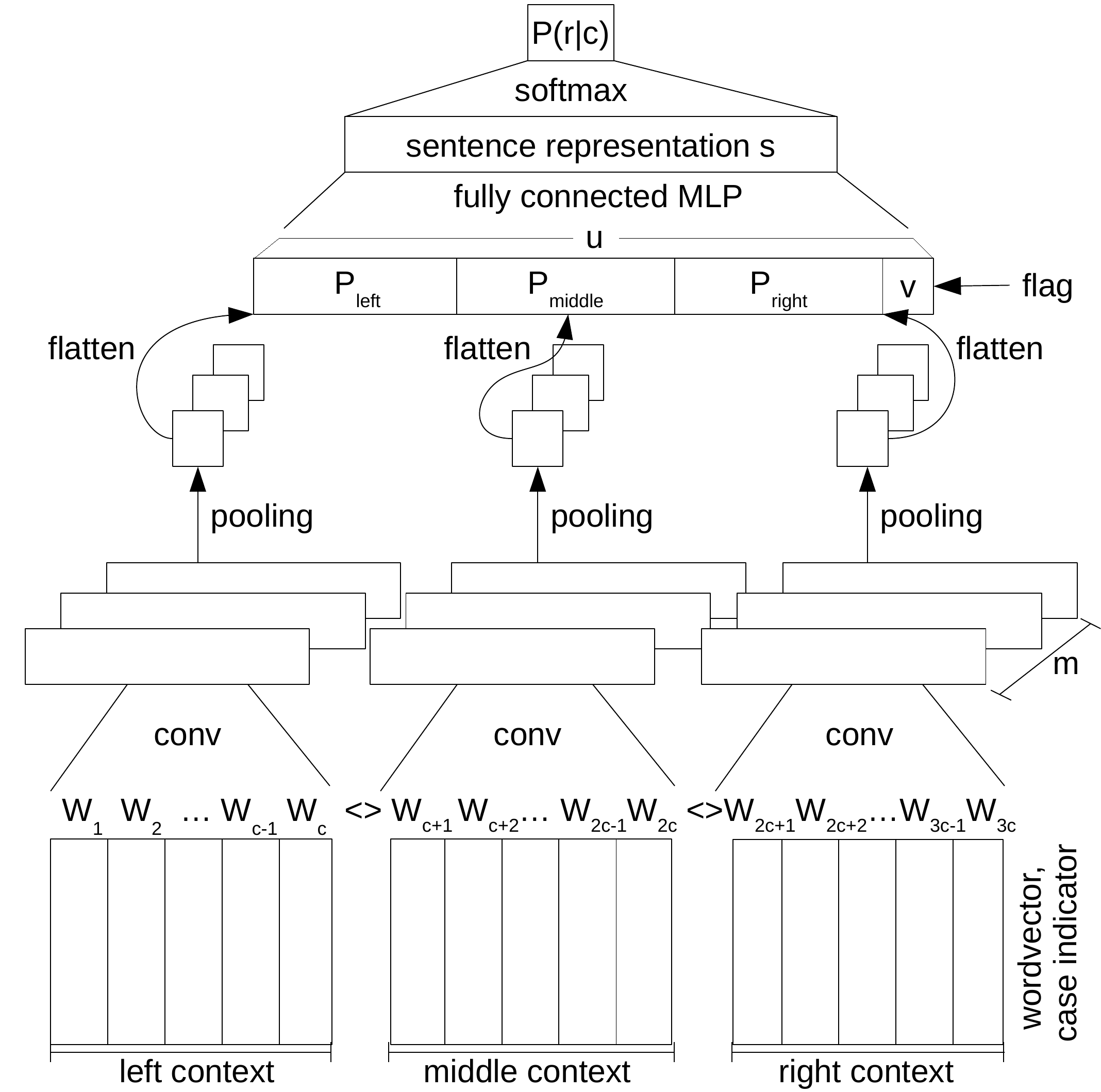}
\caption{Convolutional neural network for slot filling.}
\label{cnnFig}
\end{figure}

Figure \ref{cnnFig} depicts the structure of the CNN.

\subsection{Type-aware Convolutional Neural Networks}
\label{sec:types}
The relation arguments of the input for the binary models all have
named entity types corresponding to the expected types of the slots.
The model for the relation \texttt{per:date\_of\_birth}, for example,
only gets sentences with one relation argument being a \texttt{PERSON}
and the other relation argument being a \texttt{DATE}.
In our training data, this is ensured by the design of the extraction process
of positive and negative examples (see Section \ref{sec:trainingdata}).
In the slot filling pipeline, the candidate extraction component
extracts filler candidates based on their named entity types.
In contrast to the input of the binary models, the input
of the multi-class model can contain relation arguments
of all available types. This complicates the relation classification
task of the model.
A context for the relation \texttt{per:date\_of\_birth}, for example,
might be similar to a context for the relation \texttt{per:location\_of\_birth}.
To simplify the classification, we propose to provide the models
with the named entity types of the relation arguments.
In particular, we investigate three different settings for augmenting
the input of the multi-class model with named entity types. They are 
described in the following paragraphs.
For all settings, we use the same coarse-grained types
as we process in the slot filling pipeline, namely \texttt{PERSON,
ORGANIZATION, LOCATION, DATE, NUMBER, O}. 

\subsubsection{Pipeline Approach}
\label{sec:pipeline}
The first approach we explore is a pipeline approach
based on the slot to evaluate.
We use two binary (one-hot) type vectors $t_1$ and $t_2$ of the size
of the type vocabulary as additional input to the network
from Figure \ref{cnnFig}.
For the slot \texttt{per:employee\_or\_member\_of},
for example, the type vector $t_1$ for the first relation argument would 
consist of only one 1 at the position of \texttt{PERSON} and 0 otherwise. 
The type vector $t_2$
for the second relation argument would consist of a 1 at the position
of \texttt{ORGANIZATION} and a 1 at the position of \texttt{LOCATION}
since a person can  be employed by an organization or by a
geo-political entity.
Since the type vectors are based on the slots and the
filler extraction with named entity recognition 
(see Section \ref{component:candidateExtraction}), we call
this approach ``pipeline''.
The type vectors are then fed into a hidden layer which creates type embeddings $E_1$ and $E_2$:

\begin{equation}
 E_i = \tanh(V^\top t_i + c)
\end{equation}
with $V$ being the weight matrix and $c$ the bias term of the hidden layer.

\begin{figure}[h]
 \centering
 \includegraphics[width=.6\textwidth]{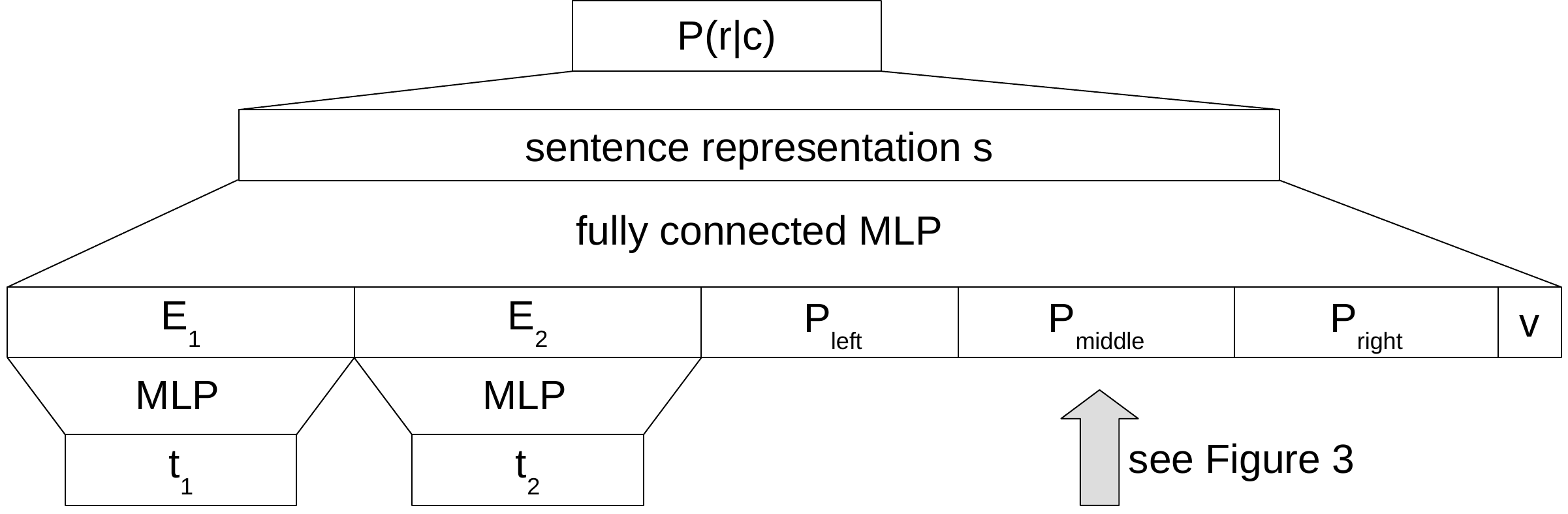}
 \caption{Integration of entity type information into multi-class CNN.}
 \label{fig:combination} 
\end{figure}

Then, we concatenate the type embeddings $E_1$ and $E_2$ with the 
pooling results of the CNN for relation classification to calculate
a type-aware sentence representation $s$.
Thus, Equation \ref{eq:hidden} becomes:
\begin{equation}
  s = \tanh(W_1^\top P_{\text{left}} + W_2^\top P_{\text{middle}} + W_3^\top P_{\text{right}} + W_4^\top v  + W_5^\top E_1 + W_6^\top E_2 + d) 
 \label{eq:hidden2}
\end{equation}
This is depicted in Figure \ref{fig:combination}.

\subsubsection{Joint Training}
\label{sec:jointTraining}
\begin{figure}[h]
\centering
\includegraphics[width=.6\textwidth]{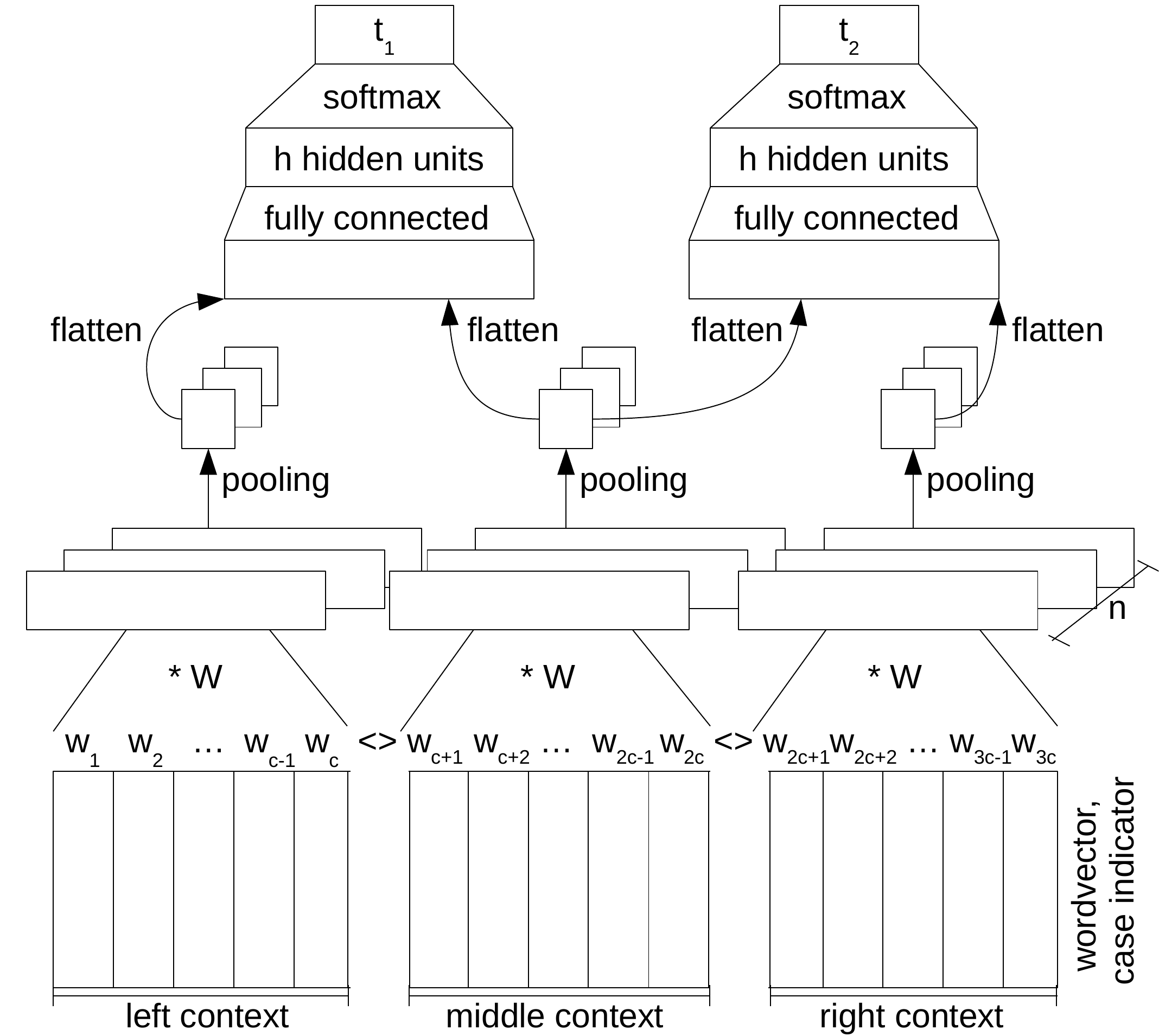}
\caption{Convolutional neural network for entity type classification; $t_1$ 
and $t_2$ are the predictions of the types of the first and second
relation argument, respectively.}
\label{fig:typePrediction} 
\end{figure}

Instead of using prior knowledge about the slots and the expected types
of their arguments, 
we propose to jointly learn entity and relation classification.
Following \citeA{EACLnoiseMitigation}, we use a convolutional
neural network to predict scores for the different types (see Figure \ref{fig:typePrediction}).
We treat this as a multi-label classification task and use the sigmoid function to ensure that 
the scores for each class are between 0 and 1.
We then use the scores as type vectors $t_1$ and $t_2$
in Figure \ref{fig:combination}. This is similar to the architecture
\textsc{Predicted-Hidden} from \citeA{EACLnoiseMitigation} with the following
differences: 
We do not use entity embeddings for modeling
the relation arguments (see Section \ref{sec:neuralNets-remarks} for our motivation)
and also integrate the flag for the order of the relation
arguments. This flag is highly relevant for slot filling since there is an
inverse slot for almost all slots.

For jointly training the CNN for entity classification and
the CNN for relation classification, we use the following loss function:
\begin{equation}
 L = (1-\alpha) \cdot L_{\text{rel}} + \frac{\alpha}{2} \cdot L_{\text{type1}} + \frac{\alpha}{2} \cdot L_{\text{type2}}
\end{equation}
The weight $\alpha$ controls the ratio between the relation
classification loss $L_{\text{rel}}$ and the losses of entity type classification $L_{\text{type1}}$
and $L_{\text{type2}}$. It is tuned on the development set.

\subsubsection{Structured Prediction}
\label{sec:structuredPrediction}

\begin{figure}[h]
\centering
\includegraphics[width=.9\textwidth]{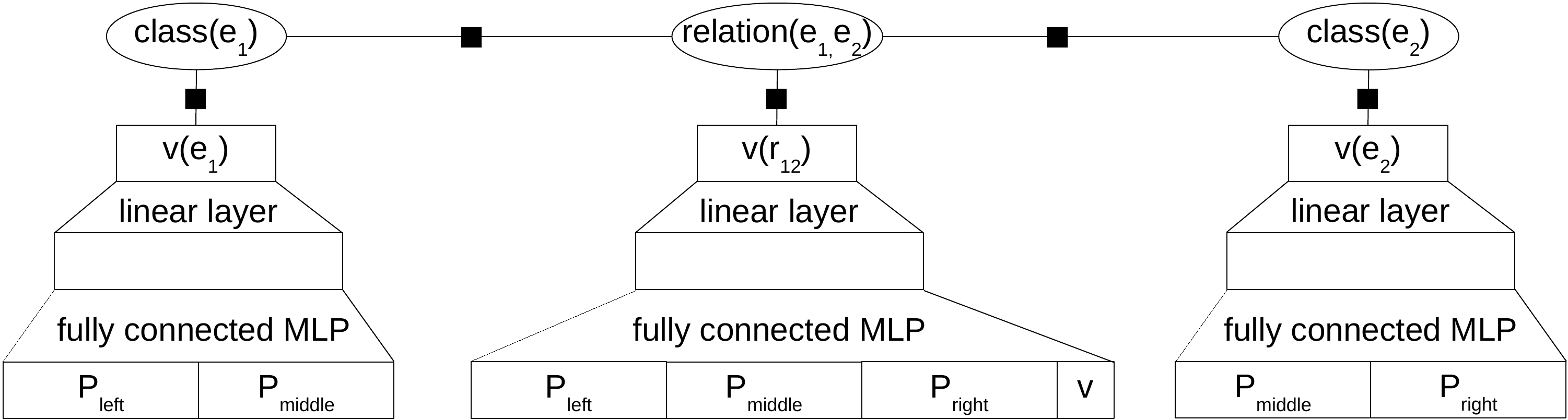}
\caption{CNN with structured prediction for type-aware slot filling relation classification.}
\label{fig:global}
\end{figure}

The third approach for integrating entity information into a 
convolutional neural network for relation classification
is based on structured prediction, 
as we originally presented for a table filling evaluation of 
entity and relation recognition \cite{adelEmnlp2017}.
While we applied it only to a manually labeled dataset in that previous work,
we now adopt it to the slot filling pipeline setting with distantly 
supervised training data for the first time.
Figure \ref{fig:global} shows the architecture of the model.
Again, the context is split into three parts: left, middle and right context.
For each context, a convolutional and 3-max pooling layer computes a 
representation with weights shared across contexts.
For predicting the class of the first entity, we use the representation
of the left and middle context; for predicting the class of the second
entity, we use the representation of the middle and right context.
To calculate scores for the possible relations, all three contexts are used.
In contrast to the model we proposed earlier \cite{adelEmnlp2017}, we again
use the flag for the order of the relation arguments for classifying the relation.
Also, we do not compute representations
for entities for slot filling, 
as motivated in Section \ref{sec:neuralNets-remarks}.
The scores of the two entity classes and the relation class are then
fed into a conditional random field (CRF) output layer which optimizes
the following sequence of predictions:
\begin{equation*}
[\text{class of } e_1, \text{relation } r_{12} \text{ between } e_1 \text{ and } e_2, \text{class of } e_2]
\end{equation*}

In particular, we apply a linear-chain conditional random field.
Thus, the model learns scores for transitions $T$
between the class of the first entity and the relation and 
between the relation and the class of the second entity.
As a result, it approximates the joint probability
of entity classes $C_{e_1}$, $C_{e_2}$ and relations
$R_{e_1e_2}$ as follows:

\begin{equation}
P(C_{e_1}, R_{e_1e_2}, C_{e_2}) \approx 
P(C_{e_1}) \cdot P(R_{e_1e_2}|C_{e_1}) \cdot P(C_{e_2}|R_{e_1e_2})
\end{equation}
Our intuition behind this is that the dependency between relations
and entities is stronger than the dependency between
the two entities.

Given neural network activations $v$ for the different entity
and relation classes, the input sequence $d$ to the CRF layer is
\begin{equation}
 d = [v(e_1), v(r_{12}), v(e_2)]
 \label{eq:inputSeq}
\end{equation}
This sequence is padded with a begin and end tag and
used to compute the score for a particular sequence $s$ in the following way:
\begin{equation}
 \text{score}(s) = \sum_{i = 0}^{n}T_{s_is_{i+1}} + \sum_{i=1}^nd_{is_i}
 \label{eq:score}
\end{equation}
with $T$ being the transition scores (randomly initialized and learned
during training) and $d$ storing the neural network activations (see Equation \ref{eq:inputSeq}).
Following \shortciteA{crfCode}, we assume that all variables
live in log space and, therefore, use the sum in Equation \ref{eq:score}.

For training, we normalize the score of the gold sequence over the scores
of all possible sequences. We compute all possible sequences with the forward algorithm.
To compute the best path during testing and get probabilities for the different relation
classes, we apply the viterbi and forward-backward algorithm, respectively \cite{hmmPaper}.

\section{Experiments and Results}
\label{sec:results}
In this section, we describe our datasets (Section \ref{sec:data}),
and present and discuss our results.
We conduct two sets of experiments: First, we evaluate
our models in a pure relation classification setup (Section \ref{sec:resultsBenchmark}).
Second, we show the performance of the slot filling pipeline when
using our models in the slot filler classification component (Section \ref{sec:end-to-end}).
Section \ref{sec:discussion} discusses the difference between the results presented in 
Section \ref{sec:resultsBenchmark} and Section \ref{sec:end-to-end}.
Finally, Section \ref{sec:soaSF} sets our results in the context of other state-of-the-art slot filling pipelines.

\subsection{Data}
\label{sec:data}
This subsection describes the different datasets we created for our experiments:
Section \ref{sec:trainingdata} presents the training data,
Section \ref{sec:benchmark} reviews the slot filling relation classification
benchmarks we use to optimize our models (development set) and to evaluate their
performance outside of the slot filling pipeline (test set).
Section \ref{sec:dataMulti} describes how we transform the training
data which has been created for binary models into a dataset for training
multi-class models.

\subsubsection{Training Data}
\label{sec:trainingdata}
The slot filling shared task does not offer a training dataset
for relation classification models. Therefore, it is necessary to create one.
Since manual labeling is expensive and does not scale to large
amounts of data, we choose a distantly supervised labeling approach \shortcite{distant}.
In particular, we create a large set of 
training examples using distant supervision
over Freebase relation instances \cite{Freebase} and the following corpora:
\begin{itemize}
 \item TAC source corpus (LDC2013E45)
 \item NYT corpus (LDC2008T19)
 \item subset of ClueWeb (\citeauthor{clueweburl}-URL)
 \item Wikipedia
 \item Freebase description fields
\end{itemize}
Negative examples for each relation are created by 
extracting sentences with entity
pairs with the correct named entity tags for the given slot but without the given
relation according to Freebase. 
However, as mentioned before, Freebase is incomplete. Thus, if a relation 
between two entities is not stored in Freebase, it
does not mean that it does not exist in reality. Therefore, we
clean the negative examples with
trigger words and patterns: If a trigger/pattern of the given relation appears in the
sentence, we do not include it in the set of negative examples.
The list of triggers has been compiled manually based on the
official slot descriptions and examples provided by 
TAC (\citeauthor{slotsurl}-URL).
It consists of a few high-precision patterns for each slot.
By manually investigating random subsets of the filtered examples,
we find the pattern set to be very effective in reducing the number 
of false negative labels.

\begin{figure}[h]
\centering
 \includegraphics[width=.5\textwidth]{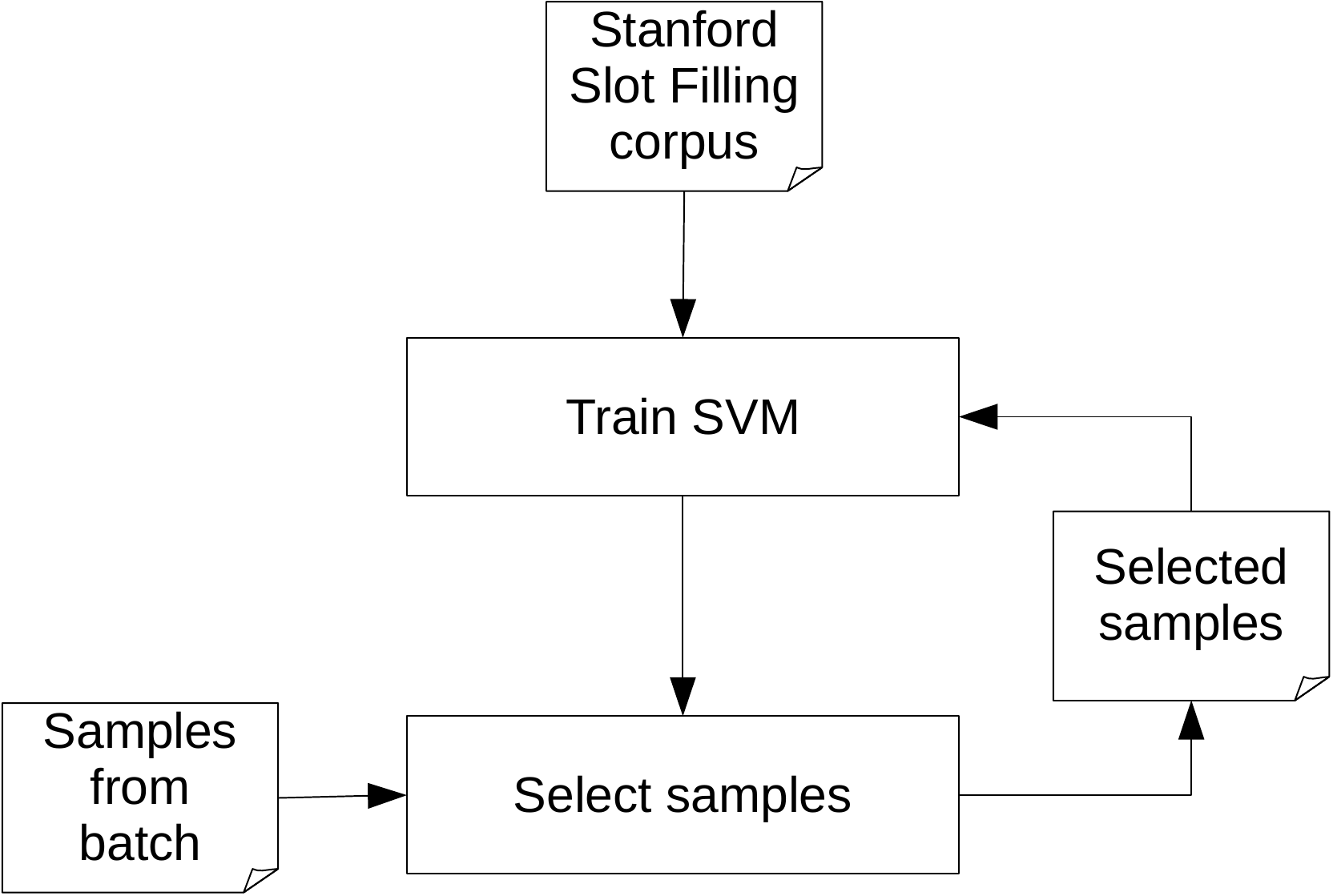}
 \caption{Training data selection process.}
 \label{dataSelection}
\end{figure}

To reduce the number of false positive labels (and further improve 
the negative labels), we perform an 
automatic training data selection process.
First, the extracted training samples are divided into $k$ batches.
Then, we train one SVM per slot on the annotated slot filling dataset 
released by \shortciteA{active}.
Thus, the classifiers are trained on data with presumably correct
labels and should, therefore, be able to help in the process
of selecting additional data.
For each batch of training samples, we use the classifiers to
predict labels for the samples and select those samples for which the
distantly supervised label corresponds to the predicted label with a 
high confidence of the classifier.
Those samples are, then, added to the training data and
the SVMs are retrained to predict the labels for the next batch.
This process is depicted in Figure \ref{dataSelection}.

\subsubsection{Development Data: Slot Filling Benchmark}
\label{sec:benchmark}
To optimize the parameters and to test the slot filling relation
classifiers outside of the slot filling pipeline, we build a slot filling relation classification
dataset, leveraging
the existing manually labeled system outputs from the previous slot
filling evaluations \cite{adelNaacl2016}: We extract the supporting 
sentences from the system outputs
and automatically determine the position of the entity and the filler.
Then, we label each sentence as correct or wrong according to the
manual assessment provided by the shared task organizers.
Due to differences in the offset
calculation of some systems, we cannot use all available data:
We extract 39,386 relation classification instances out of the 59,755 system output
instances which have been annotated as either
completely correct or completely incorrect by the shared task organizers.
Thus, the resulting
dataset has a reasonable number of examples 
with presumably clean labels.
For our experiments, we split the data into a development part (data from
slot filling evaluations 2012-2013) and a test part (data from
slot filling evaluations 2014). We tune the parameters of our models
on the development part and use the test part as a first indicator of their performance
on unseen data.
For more details on the data 
and a script to reproduce the data, see \citeA{adelNaacl2016}.

\subsubsection{Data for Multi-class Models}
\label{sec:dataMulti}
Since our training dataset has been created for binary models, the 
negative examples for each slot need to be processed for 
the multi-class setting: For example, a negative instance for the slot \texttt{per:date\_of\_birth}
is not automatically a negative instance for the slot \texttt{per:date\_of\_death}.
Therefore, we filter the negative instances in the
training data with pattern lists: A negative instance
that includes a trigger for any of our positive slots is deleted from the set.
The remaining negative instances are labeled with an artificial class \texttt{N}.
Finally, we found it beneficial on the slot filling relation classification
benchmark to subsample the number of negative instances. Thus,
we use the same number of negative instances as non-negative instances.

Note that we only modify the training set and still 
use the original development and test sets from the slot filling relation
classification benchmark for our experiments in order to
compare the multi-class models with the binary models.

\subsection{Results on Slot Filling Benchmark}
\label{sec:resultsBenchmark}

In this subsection, we present the performance of our models on the slot filling benchmark dataset (which
is described in Section \ref{sec:benchmark}).
Section \ref{sec:baselines} summarizes the baseline models to which we compare our type-aware 
convolutional neural networks, and Section \ref{sec:evaluationBenchmark} describes
the evaluation measure. Finally, Section \ref{sec:benchmarkResults} provides and 
discusses the performance of our different models.

\subsubsection{Baselines and Support Vector Machines}
\label{sec:baselines}
In earlier work \cite{adelNaacl2016}, we compared various 
models on the slot filling relation classification benchmark dataset
described in Section \ref{sec:benchmark}.
In this paper, we use the same baseline models, namely \textsc{Mintz++} \cite{distant}
and \textsc{MimleRe} \shortcite{mimlre}. For another comparison, we also train multi-class
variants of the SVM models from \citeA{adelNaacl2016}.
For training the multi-class SVM, we apply
LinearSVC from scikit 
learn (\citeauthor{svmurl}-URL)
with automatically adjusted
class weights and compare the one-vs-rest training strategy
(which actually trains binary classifiers by contrasting the 
examples from one class against the examples from all other classes)
with the multi-class training strategy by \citeA{crammersinger} (which
changes the objective function to optimizing multiple classes jointly).
The SVM is implemented using liblinear 
\shortcite{liblinear}.

\subsubsection{Evaluation}
\label{sec:evaluationBenchmark}
The different models are evaluated using $F_1$, the harmonic mean of precision $P$ and recall $R$ of the classifiers:
\begin{equation}
F_1 = \frac{2 \cdot P \cdot R}{P + R}
\end{equation}
We calculate both slot-wise $F_1$ scores and a macro $F_1$ score which is the average of the scores over all slots.

  \begin{table}[h]
 \small
 \centering
 \begin{tabular}{l|cc|ccc|ccccc}
 & \textsc{Mintz} & \textsc{Miml} & \multicolumn{3}{c|}{SVM} & \multicolumn{5}{c}{CNN}\\
 &  &  & binary & o-v-r & c-s & binary & multi & +p & +j & +s\\
 \hline
per:age & .71 & .73 & .74 & .70 & .70 & \textbf{.76} & .68 & .72 & .66 & .67\\
per:alternate\_names & .03 & .03 & .02 & .00 & .00 & \textbf{.04} & .00 & .00 & .00 & .00\\
per:children & .43 & .48 & \textbf{.68} & .39 & .45 & .61 & .48 & .44 & .44 & .36\\
per:cause\_of\_death & .42 & .36 & .32 & .00 & .00 & \textbf{.52} & .11 & .29 & .00 & .06\\
per:date\_of\_birth & .60 & .60 & .67 & .33 & .57 & .77 & \textbf{.80} & \textbf{.80} & .73 & .73\\
per:date\_of\_death & .45 & .45 & .54 & .60 & \textbf{.62} & .48 & .51 & .46 & .39 & .59\\
per:empl\_memb\_of & .36 & \textbf{.37} & .36 & .21 & .15 & \textbf{.37} & .28 & .29 & .25 & .28\\
per:location\_of\_birth & .22 & .22 & .27 & .31 & .33 & .23 & \textbf{.36} & .20 & .34 & .35\\
per:loc\_of\_death & .41 & \textbf{.43} & .34 & .32 & .35 & .28 & .28 & .19 & .25 & .21\\
per:loc\_of\_residence & .11 & .18 & \textbf{.33} & .11 & .08 & .23 & .15 & .06 & .22 & .25\\
per:origin & .48 & .46 & \textbf{.64} & .04 & .02 & .39 & .11 & .30 & .13 & .17\\
per:schools\_att & \textbf{.78} & .75 & .71 & .58 & .62 & .55 & .45 & .47 & .56 & .68\\
per:siblings & .59 & .59 & .68 & .71 & .68 & .70 & \textbf{.73} & .54 & .63 & .68\\
per:spouse & .23 & .27 & .32 & .45 & \textbf{.49} & .30 & .39 & \textbf{.49} & .36 & .30\\
per:title & .39 & .40 & .48 & \textbf{.51} & .45 & .46 & .42 & .43 & .44 & .48\\
org:alternate\_names & .46 & .48 & .62 & .59 & .52 & \textbf{.66} & .58 & .55 & .50 & .58\\
org:date\_founded & .71 & .73 & .70 & .60 & .60 & .71 & .63 & .65 & \textbf{.74} & .69\\
org:founded\_by & .62 & .65 & \textbf{.74} & .70 & .68 & .68 & .71 & .43 & \textbf{.74} & .73\\
org:loc\_of\_headqu & .19 & .20 & .42 & .14 & .11 & \textbf{.45} & .24 & .42 & .21 & .34\\
org:members & .06 & .16 & .13 & .15 & \textbf{.31} & .04 & .17 & .07 & .17 & .11\\
org:parents & .14 & .17 & \textbf{.20} & .10 & .11 & .16 & .14 & .09 & .12 & .10\\
org:top\_memb\_empl & .44 & .46 & .55 & .54 & .51 & .53 & .55 & .49 & \textbf{.58} & \textbf{.58}\\
\hline 
macro $F_1$ & .40 & .42 & \textbf{.48} & .37 & .38 & .45 & .40 & .38 & .38 & .41
 \end{tabular}
    \caption{$F_1$ results on slot filling benchmark test data (from 2014). The columns
show our binary and multi-class SVMs and CNNs as well as our type-aware CNN models
in comparison to two standard baseline models for relation classification: \textsc{Mintz} and \textsc{Miml}
(see Section \ref{sec:baselines} for more information).
    o-v-r: one-vs-rest; c-s: Crammer and Singer; p: pipeline (Section \ref{sec:pipeline}), 
j: joint training (Section \ref{sec:jointTraining}), s: structured prediction (Section \ref{sec:structuredPrediction}).}
 \label{tab:yearwiseResults}
 \end{table}

\subsubsection{Results}
\label{sec:benchmarkResults}
Table \ref{tab:yearwiseResults} provides slot-wise results for 
the baseline models, the SVMs as well as for the different 
CNN setups: binary CNNs,
a multi-class CNN without entity type information,
a multi-class CNN with slot-based entity types (+p),
a multi-class CNN with entity type probabilities jointly trained
with the relation classification CNN (+j), and
a multi-class CNN with a structured prediction (CRF) output layer which optimizes
a sequence of entity and relation classes (+s).
Note that Table \ref{tab:yearwiseResults} shows results for 22 slot types instead
of 65 slot types mentioned in Section \ref{sec:task}. The reason is that
we merged all location slots and all inverse slot pairs into one slot type (see Section \ref{sec:neuralNets-remarks}). 
For example, our slot type
\texttt{per:children} actually covers both original slot types \texttt{per:children} and \texttt{per:parents},
and our slot type \texttt{org:founded\_by} covers even four slot types, namely \texttt{org:founded\_by},
\texttt{per:organizations\_founded}, \texttt{org:organizations\_founded} and 
\texttt{gpe:organizations\_\-fo\-und\-ed}. As a result, our merged slot types in 
Table \ref{tab:yearwiseResults} actually cover 54 out of all 65 slot types.
We provide an overview which original slot types we cover with our models 
in Table \ref{tab:coveredslots} in the appendix. 
For the remaining slot types, we were not able to extract enough training data with distant supervision to train
machine learning models. For those slots, our slot filling pipeline falls back to pattern matching.

In general, the binary
models perform better than the multi-class models, even
when adding entity type information to the latter.
For example, the binary CNN achieves better results than the multi-class CNN
without entity type information for 13 out of 22 slot types
and better results than any multi-class CNN (with or without entity type information)
for nine out of 22 slot types.
We assume that one reason might be that the convolutional
filters of the binary models can concentrate on learning features
for just one relation type while the filters of the multi-class models
need to learn features which can be used to discriminate all relation types.
All binary models outperform the baseline models \textsc{Mintz} and \textsc{Miml}.
The binary SVM performs slightly better than the binary CNN.
In contrast to these results,
the multi-class CNN outperforms the multi-class SVM. 
However, both models seem to struggle with the large
number of output classes.

While the model with jointly trained entity classification achieves
slightly better results on the development set than the multi-class model without named 
entity information (0.53 vs.\ 0.52), 
this improvement is not transferred to the test set.
The structured prediction model generalizes better to an unseen test set
than the other type-aware models and comes closer to the results of the binary CNN.
Slots for which entity type information seems to help the most
are \texttt{per:date\_of\_birth} and \texttt{per:location\_of\_birth},
two slots with similar contexts. Similarly, type-aware CNNs achieve the best
results for \texttt{org:date\_founded} and \texttt{org:founded\_by}.
Despite their lower results on average, the multi-class models 
generalize better to unseen test data than the binary models for some slots, such 
as \texttt{per:date\_of\_death}, \texttt{per:schools\_attended} or \texttt{per:spouse}.

In the next section, we provide results for using the different models
in the slot filling pipeline.

\subsection{Slot Filling Pipeline Results}
\label{sec:end-to-end}

In this subsection, we present the performance of our slot filling pipeline 
(described in detail in Section \ref{sec:overview}) on the official slot filling evaluation data
from 2015. It consists of 1951 queries with manual assessments of the outputs of the systems
that were submitted to the shared task evaluations. 930 of those queries are multi-hop queries,
i.e., they require the system to first fill one slot and then fill a second slot based on the answers
of the first slot (see Section \ref{sec:task}).
Section \ref{sec:systems} presents the different system configurations we evaluate.
Section \ref{sec:evaluationPipeline} describes the evaluation measures and
Section \ref{sec:resultsPipeline} provides and discusses the results.

\subsubsection{System Configurations}
\label{sec:systems}
In this experiment, we run the whole slot filling pipeline as described in Section \ref{sec:overview}.
The different configurations we evaluate differ from one another in terms of the relation classification
models that are used in the slot filler classification component. Note that all configurations
use the pattern matching module in addition to the machine learning models.
The numbers we provide in the following to distinguish the different configurations 
correspond to the numbers in Table \ref{tab:results-ldcMax2}.
System I only uses binary SVMs in combination with patterns. Thus, it can be used to assess the impact of CNNs in general.
System II combines binary SVMs and binary CNNs.
To assess the impact of joint training with entity types also for the binary models, we add another
configuration, system III, that uses binary SVMs and binary CNN models which are jointly trained 
on entity typing and relation classification. Thus, in contrast to system II, it has the ability 
of correcting wrong entity types from the candidate extraction component of the pipeline.
System IV uses multi-class SVMs instead of binary SVMs as in system I.
For the multi-class SVMs, we use the one-vs-rest training strategy
since this led to slightly better results on the development part
of the slot filling relation classification benchmark (0.51 vs.\ 0.50).
System V combines multi-class SVMs with multi-class CNNs. Thus, comparing system II to
system V allows us to assess the performance difference of binary and multi-class models for slot filling.
Finally, systems VI, VII and VIII combine multi-class SVMs with type-aware multi-class CNNs,
our main contribution in this paper. System VI integrates pipeline-based CNNs, as described in Section \ref{sec:pipeline},
into the slot filling pipeline.
In system VII, we apply CNNs that have been jointly trained on entity and relation classification,
as presented in Section \ref{sec:jointTraining}. The last configuration, system VIII, uses
CNN models trained with structured prediction of both entity and relation classes.
Note that we only compare pure binary to pure multi-class classification modules in this paper.
We also experimented with combining binary SVMs with
multi-class CNNs but did not obtain additional performance gains with this setup.

\subsubsection{Evaluation}
\label{sec:evaluationPipeline}
The slot filling pipeline is evaluated using the official measures from the shared task.
In particular, we report two measures, one based on the micro $F_1$ score (the $F_1$ score over all examples,
giving slots with more examples higher weight),
and another one based on the macro $F_1$ score (the average of slot-wise $F_1$ scores, weighting all slots equally):
``CSLDC max micro'' is the micro precision,
recall and $F_1$ score over all queries (thus, ``micro''). If the hop 0
sub-query occurred several times in the query set (with different hop 1 sub-queries), that answer
to the hop 0 sub-query
is scored which leads to the maximum results over both
hops (thus, ``max'').
``CSLDC max macro'' is the average $F_1$ score over all slots (thus, ``macro'').
All scores in this section are calculated using the official shared task 
scoring scripts. Their readme file provides more details on the scoring procedure (\citeauthor{scoringurl}-URL).

\subsubsection{Results}
\label{sec:resultsPipeline}
Table \ref{tab:results-ldcMax2} provides the scores
of our slot filling system with our newly introduced classification models
in comparison to SVMs and CNNs (in both binary and multi-class variants)
without entity type information.

\begin{table}[h]
\centering
\begin{tabular}{llll|rrr|r}
& & & & \multicolumn{3}{c|}{micro} & macro \\
&  & SVM & CNN & $P$ & $R$ & $F_1$ & $F_1$\\
\hline
\multirow{8}{*}{\rotatebox{90}{hop 0}}
& (I) & binary & - & 29.12 & 26.18 & 27.57 & 32.50\\
& (II) & binary & binary & 31.79 & 28.23 & 29.91 & 34.20\\
& (III) & binary & binary + j & 30.11 & 28.23 & 29.14 & \textbf{34.81}\\
\cline{2-8}
 & (IV) & multi & - & 25.25  & 12.07 & 16.33 & 18.44\\
& (V)  & multi & multi & \textbf{34.42} & 26.66 & 30.04 & 32.82\\
& (VI) & multi & multi + p & 23.58 & \textbf{28.55} & 25.83 & 30.91\\
 & (VII) & multi & multi + j & 32.42 & 27.84 & 29.95 & 33.14\\
& (VIII) & multi & multi + s & 33.33 & 27.68 & \textbf{30.25} & 33.98\\
\hline\hline
\multirow{8}{*}{\rotatebox{90}{hop 1}}
& (I) & binary & - & 7.36 & 4.78 & 5.80 & 6.13\\
& (II) & binary & binary & 9.80 & \textbf{7.00} & \textbf{8.17} & 8.28\\
& (III) & binary & binary + j & 8.90 & 6.56 & 7.55 & 8.73 \\
\cline{2-8}
&(IV) & multi & - & 7.11 & 3.67 & 4.84 & 4.34\\
& (V) & multi & multi &  12.59 & 3.89 & 5.94 & 7.78\\
& (VI)  & multi & multi + p & 6.62 & 3.00 & 4.13 & 4.66\\
& (VII) & multi & multi + j & \textbf{13.47} & 5.00 & 7.29 & 8.15\\
& (VIII) & multi & multi + s & 12.24 & 5.22 & 7.32 & \textbf{9.24}\\
\hline\hline
\multirow{8}{*}{\rotatebox{90}{all}}
& (I) & binary & - &  21.75 & 17.30 & 19.27 & 23.06\\
& (II) & binary & binary & 23.80 & \textbf{19.42}  & 21.39 & 24.92\\
& (III) & binary & binary + j & 22.52 & 19.23 & 20.75 & \textbf{25.47}\\
\cline{2-8}
& (IV) & multi & - &  17.38 & 8.58 & 11.49 & 13.39\\
& (V) & multi & multi &  \textbf{29.60} & 17.20 & 21.76 & 23.86\\
 & (VI) & multi & multi + p & 20.02 & 17.94 & 18.92 & 21.51\\
& (VII) & multi & multi + j & 27.97 & 18.36 & \textbf{22.17} & 24.20\\
& (VIII) & multi & multi + s & 27.70 & 18.36 & 22.08 & 25.12
\end{tabular}
\caption{Slot filling pipeline results for our different models and model combinations. Evaluation measure: CSLDC max micro/macro.
p: pipeline, j: joint training, s: structured prediction as in Table \ref{tab:yearwiseResults}.}
\label{tab:results-ldcMax2}
\end{table}

In contrast to the slot filling relation classification results (see Table 
\ref{tab:yearwiseResults}), most of the multi-class models (systems IV--VIII in Table \ref{tab:results-ldcMax2}) perform comparable
or even better than the binary models (systems I--III). The best micro $F_1$ results for hop 0
and all (both hops) are achieved by using multi-class
classification models (systems V,VII). 
The multi-class models have higher
precision than the binary models across all hops 
(hop 0, hop 1 and all).
The multi-class models with joint training of entity types and relation
classification (system VII) achieves the highest overall micro $F_1$ score.
In terms of macro $F_1$, the binary model with jointly learned types (system III) and the multi-class 
model with structured prediction (system VIII) perform best.
This suggests that the opportunity to correct wrong entity types from the candidate extraction component
is crucial and that joint modeling with entity types especially improves the performance on rare relations.

\subsection{Discussion of Results: Benchmark vs.\ Pipeline}
\label{sec:discussion}

The performance ranking of the models on
the benchmark dataset for slot filling relation classification (Table 
\ref{tab:yearwiseResults}) is different to the performance ranking in the
slot filling pipeline evaluation (Table \ref{tab:results-ldcMax2}).
This seems to be in contradiction to the positive correlation
of results we reported earlier \cite{adelNaacl2016}. We assume that the reason
is domain mismatch. While the positive correlation
was calculated for running the pipeline on 2013/2014 slot filling
evaluation data \cite{adelNaacl2016}, the pipeline is now run on 2015 evaluation data.
In 2015, the evaluation corpus for slot filling was changed,
introducing many more discussion forum documents
and significantly changing the ratio of domains (see Table \ref{tab:domains}).
This leads to a severe domain mismatch challenge for the components of the slot filling
pipeline and reduces the correlation with the benchmark dataset which has been
built based on 2012-2014 data. Nevertheless, participants of the slot filling
shared task only have previous evaluation data available for developing and
tuning their models. 
Therefore, we argue that it is still important to also evaluate
models on the slot filling
relation classification benchmark.
\begin{table}[h]
\centering
\begin{tabular}{l|rr|rr}
 & \multicolumn{2}{c|}{until 2014} & \multicolumn{2}{c}{2015}\\
 & number & ratio & number & ratio\\
 \hline 
news documents & 1,000,257 & 47.65\% & 8,938 & 18.19\%\\
web documents & 999,999 & 47.63\% & 0 & 0\%\\
discussion forum & 99,063 & 4.19\% & 40,186 & 81.81\%\\
\end{tabular}
\caption{Domains in slot filling corpora.}
 \label{tab:domains}
\end{table}

\subsection{Comparison with State of the Art}
\label{sec:soaSF}
Finally, we set our results in the context of state of the art
on the 2015 evaluation dataset for slot filling. 

\begin{table}[h]
\centering
 \begin{tabular}{l|l|r|c}
	 rank & team & micro $F_1$ & distant supervision?\\
  \hline 
	 1 & Stanford  & 31.06 & no\\
	 \hline
	 2 & UGhent  & 22.38 & yes\\
	 & CIS (with type-aware CNN) & 22.17 & yes\\
	 3 & CIS (our official submission) & 21.21 & yes\\
	 4 & UMass  & 17.20 & yes\\
	 5 & UWashington & 16.44 & yes\\
	 \hline 
	 median & - & 15.32 & - \\
 \end{tabular}
\caption{State-of-the-art results for slot filling.
Stanford: \cite{stanford2015}, 
UGhent: \cite{ughent2015}, CIS: \cite{cis2015}, UMass: \cite{umass2015}, UWashington: \cite{uw2015}.
The third line (``CIS with type-aware CNN'') is our best result from Table \ref{tab:results-ldcMax2}.
Our official submission (forth line) did not include type-aware models.}
\label{tab:soa}
\end{table}

Table \ref{tab:soa} shows that our official submission 
to the shared task in 2015,
which did not include type-aware models,
was ranked at third position. 
With our type-aware convolutional neural networks, we can 
improve our result. Our pipeline performs comparable to the 
second rank now. It is considerably better than the system
on rank 4 and performs clearly above median.
Only the results of the top-performing system are still superior. 
As also described in Section \ref{sec:relWork}, there
are two differences relevant to this: 
First, the top-performing system does not use information retrieval,
like our system and most other systems, but stores preprocessed versions
of the corpus in a database, including an index for all occurring entities.
This requires extensive corpus preprocessing (such as the identification
of all entities along with their positions in the documents) 
and data storage (the source corpus contains millions of documents) but
makes it possible to access the query entities directly at test time.
However, this approach is only possible with prior access to the
whole corpus and cannot be applied to changing environments.
In contrast, our retrieval-based pipeline is more flexible since
it processes only those documents relevant to the input query.
The second, and arguably more important, difference
in terms of final performance is that the training dataset of the
top-performing system
has been labeled manually via crowdsourcing. In contrast, the
datasets of the other systems are created with distant supervision.
When training our models, we observed large performance differences
depending on the quality of the training data. Therefore, we suspect
that the main reason for the superior performance of the system
by \citeA{stanford2015} is their training data.
Unfortunately, obtaining manual labels is time-consuming and
challenging, even in the context of crowdsourcing. 
An example is the extension of the knowledge base schema to new relations
or a more fine-grained distinction of existing relations
which would always require manual relabeling.
Therefore, automatic methods like distant supervision are still of high relevance.
Among the systems using automatically created training data,
our pipeline is state of the art.

\section{Analysis}
\label{sec:analysis}
In this section, we analyze the behavior of the slot filling system in more detail in order
to see which pipeline components need to be improved in the future.
Section \ref{sec:recallAnalysis} analyzes the recall of the different components
of the slot filling pipeline, showing which components are responsible for which recall loss.
Section \ref{sec:errorAnalysis} presents a manual analysis of wrong system outputs,
categorizing the errors with respect to which pipeline component is responsible for them.
Finally, Section \ref{sec:ablation} provides several ablation studies, indicating the impact
of entity linking, coreference resolution and type-aware neural networks.

\subsection{Recall Analysis}
\label{sec:recallAnalysis}
Our first analysis investigates the recall of the different components 
and is similar to the analysis by \citeA{analysisRecall}.
In particular, we evaluate the components of our system before the slot filler classification
module. Thus, we measure which recall our system could achieve with a
perfect slot filler classification module that does not lose any recall.

\begin{figure}[h]
\centering
 \includegraphics[width=.8\textwidth]{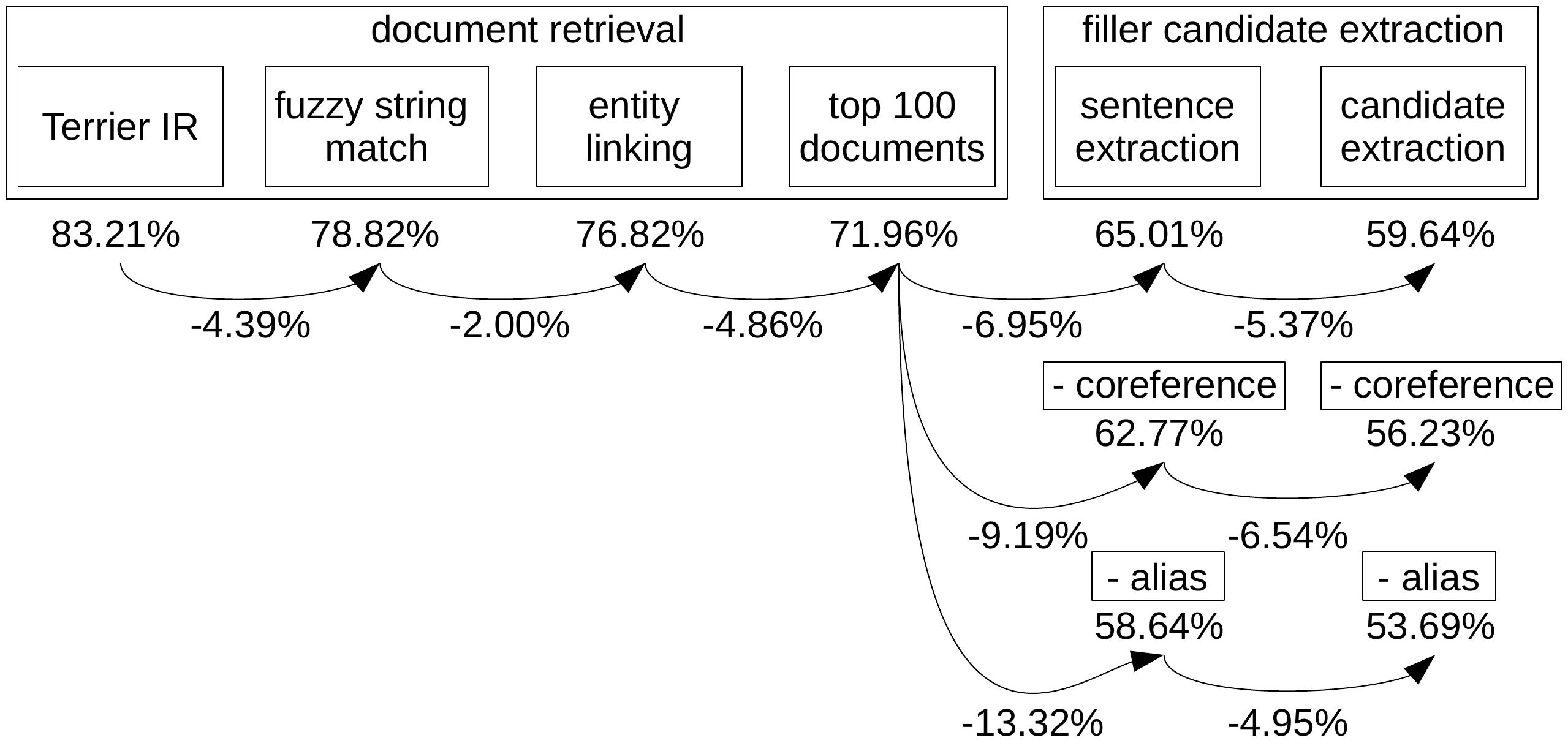}
 \caption{Analysis of recall after the application of the different pipeline components.}
 \label{fig:recallanalysis}
\end{figure}

Figure \ref{fig:recallanalysis} shows the results on the 
slot filling assessment data from 2015 for hop 0: 
Information retrieval with \textsc{Terrier} IR and
fuzzy string match
is able to achieve a recall of 78.82\%. 
The entity linking component hurts recall a bit. However, it also increases
precision which leads to better overall results 
(cf., Section \ref{sec:ablationEL}).
Evaluating only the top 100 documents instead of all extracted documents from \textsc{Terrier}
(maximum 300), leads to a recall loss of almost 5\%. Thus, allowing the slot filling
system a longer run time for processing all extracted documents could lead to a 
higher final recall (but potentially also to more false positive extractions and, thus, 
a lower precision). As mentioned before, choosing only the 100 most relevant documents
has led to the best time-performance trade-off on data from previous evaluations 
(2013 and 2014). 
The sentence extraction component extracts the relevant sentences
quite successfully with an additional recall loss of only 6.95\%. Evaluating this 
component in more detail shows the importance of coreference resolution and aliases:
The recall loss without coreference resolution is almost 10\%, the recall loss
without aliases is more than 13\%. 
Finally, the candidate extraction component
is able to extract most of the relevant candidates, yielding an overall recall of
59.64\% before slot filling relation classification. Without coreference resolution for 
sentence extraction, the overall recall is 56.23\%, without alias information
for sentence extraction, the overall recall is 53.69\%.
Assuming a perfect slot filler classification component with $P=100\%$ and
$R=100\%$, the maximum $F_1$ score of the whole slot filling system would
be 74.72\%. This number is about twice as high as the performance
of the best slot filling system 2015 \cite{stanford2015} (see Table \ref{tab:soa}) but still
low compared to other NLP tasks. This illustrates the difficulties of
the slot filling task and the importance of all individual components
of the pipeline since especially recall losses cannot be recovered
by subsequent components.

\subsection{Error Analysis}
\label{sec:errorAnalysis}
In our second analysis, we manually analyze 120 errors of our system from
the official 2015 evaluations, i.e., its wrong (false positive) predictions.
While Section \ref{sec:recallAnalysis} shows the recall loss of the different pipeline
components, this analysis categorizes which component is responsible for which
false positive prediction, and as a result, for a precision loss of the pipeline.
Table \ref{tab:errorAnalysis} shows which pipeline component
is responsible for how many errors. The numbers do not sum to 1
since for 7\% of the cases, we could not unambiguously identify
a single component as the error source.

\begin{table}[h]
\centering
 \begin{tabular}{lr}
  Error category & ratio\\
  \hline
  Alias component & 9\%\\
  Entity linking component & 2\%\\
  Candidate extraction component & 21\%\\
  Classification component & 61\%
 \end{tabular}
  \caption{Error analysis of the pipeline.}
 \label{tab:errorAnalysis}
\end{table}

The alias component especially struggles with acronyms which
can refer to several entities. An example is NL which 
is an acronym for ``National League'' in the document collection but got
wrongly recognized as an acronym for 
the query entity ``Nest Labs''.
In the candidate extraction component, most errors (16\% of 21\%)
occur in the named entity recognition part.
For example, ``Bloomberg'' is wrongly tagged as organization although
it is a person in the given context (``... people like Bloomberg ...'').
Similarly, ``Heinz'' gets tagged as person although it is
an organization in the given context 
(``MacDonald's dropping Heinz after CEO change'').
Nested named entities are also a challenge: In ``Tom Clancy games'',
for example, ``Tom Clancy'' gets tagged as a person although the whole
phrase actually forms a single entity.
For some instances (4\% of 21\%), the document has been
incorrectly split into sentences and in the remaining
cases (1\% of 21\%), coreference resolution failed.
The classification component faces a very challenging
task since most extracted filler candidates are false positives.
Thus, it has to establish precision while keeping as much
recall as possible.
Based on a manual inspection of errors, the most 
important challenge for the classification
component is long contexts which mention several relations
between several entity pairs.
An example is ``Mikhail Kalashnikov, designer of the famed Russian AK-47 assault rifle, 
died on Monday in his home city of Izhevsk, 
an industrial town 1,300 km east of Moscow, local media reported''
from which the relation \texttt{per:location\_of\_death}
between Mikhail Kalashnikov and Moscow is extracted. Thus, our classification
component correctly recognizes the relation trigger ``died [...] in'' but
assigned it to the wrong relation arguments.
This finding is in line with the study by \shortciteA{lifu2017} who also
identified long context as one of the main challenges in slot filling relation classification.

\subsection{Ablation Studies}
\label{sec:ablation}
In our last analysis, we present ablation studies showing the 
impact of entity linking, coreference resolution and type-aware neural networks. We focus on entity linking
and coreference resolution since the design choices of whether or not to integrate them into
a slot filling system are among those with the highest disagreement among
slot filling researchers (see our description of related work in Section \ref{sec:relWork}).
Furthermore, the type-aware neural networks for slot filling are the main contribution of this paper.
Table \ref{tab:ablationEL} compares the performance of our slot filling system
with and without entity linking, coreference resolution, and type-aware neural networks, respectively.

\begin{table}[h]
\centering
\begin{tabular}{llrrrr}
   &  & $P$ & $R$ & $F_1$ & $\Delta$$F_1$\\
   \hline
 \multirow{4}{*}{\rotatebox{90}{hop 0}}
 & joint & 32.42 & 27.84 & 29.95\\
 & without entity linking & 31.30 & 27.92 & 29.51 & -0.44\\
 & without coreference & 32.13 & 25.87 & 28.66 & -1.29\\
 & without CNN & 25.25  & 12.07 & 16.33 & -13.62\\
 \hline 
 \multirow{4}{*}{\rotatebox{90}{hop 1}}
 & joint & 13.47 & 5.00 & 7.29\\
 & without entity linking & 11.81 & 4.78 & 6.80 & -0.49\\
 & without coreference & 11.80 & 4.22 & 6.22 & -1.07\\
 & without CNN & 7.11 & 3.67 & 4.84 & -2.45\\
 \hline 
 \multirow{4}{*}{\rotatebox{90}{all}}
 & joint & 27.97 & 18.36 & 22.17\\
 & without entity linking & 26.56 & 18.31 & 21.68 & -0.49\\
 & without coreference & 27.25 & 16.88 & 20.85 & -1.32\\
 & without CNN & 17.38 & 8.58 & 11.49 & -10.68\\
   \end{tabular}
   \caption{Impact of entity linking and coreference resolution and type-aware convolutional
   neural networks on the slot filling pipeline.}
\label{tab:ablationEL}
\end{table}

\subsubsection{Impact of Entity Linking}
\label{sec:ablationEL}
The system performance is slightly reduced by omitting entity linking.
However, the difference of the $F_1$ scores is rather small.
This shows that the main challenges of the system lie in other
components and ambiguous names play a rather small role 
for the final results of the system.

\subsubsection{Impact of Coreference}
\label{sec:ablationCoref}
The $F_1$ score drops by 1.3 points when omitting coreference resolution.
As expected, the impact on recall is higher than the impact on precision.
However, also precision is reduced. This is because the
number of true positives is reduced considerably (from 398 to 366)
when the system does not use coreference information.
The number of false positives is also lower, but the final results show
that the impact of the number of true positives is larger.

\subsubsection{Impact of Neural Networks}
Type-aware CNNs have the largest impact on performance.
They improve both precision and recall considerably.
In contrast to SVMs, the usage of word
embeddings allows the CNNs to detect synonyms
or phrases which are similar but not the same as the
ones learned during training. Training them jointly
with entity classification allows them to
benefit from the mutual dependencies between 
entity and relation classes.

\subsection{Discussion: Lessons Learned}
When developing the slot filling system and training
the relation classification models, we had to solve several
challenges. 
In this section, we report on lessons we have learned
when developing the slot filling pipeline. Those are mainly
qualitative statements which we have found by manually
inspecting intermediate system outputs and results
in the development process. Thus, many of them are based on prior
experiments and not on concrete numbers reported in the paper.

For the slot filling pipeline, it is 
especially useful to keep the extracted filler candidates and
contexts as clean as possible, e.g., by applying genre-specific
document processing steps and manually cleaning 
filler lists for string slots. Moreover, 
it is important for the recall of the system to 
extend the integration
of coreference resolution, for instance, by applying it 
to both relation arguments. Our recall analysis and 
ablation study also show
the positive impact of coreference resolution on the overall results.
Based on our error analysis,
the candidate extraction module and the classification
module are responsible for most of the errors of the overall system.
Thus, for future work, it is essential to focus on those two components.

The performance of the relation classification models is 
mainly influenced by their
training data and the input they get from the pipeline.
If the input is very long, the models might extract wrong relations
or relations between different entities than the query entity and the
filler candidate. This is a particular challenge of slot filling since 
the quality of the inputs to the relation classification models directly depends on the
previous system components. 
The significant difference of the top-ranked slot filling system
to all the other systems in 2015 emphasizes the importance of
a high-quality training dataset for slot filling relation classification.
In an analysis of random subsamples of our training dataset before and after
reducing wrong labels with patterns and self-training, 
we saw the importance of cleaning the noisy labels from distant 
supervision. Despite our automatic cleaning steps, our dataset still includes noise and a promising future
research direction might be the exploration of techniques for further enhancing
the data or collecting new data without distant supervision.
Furthermore, our results suggest that
multi-class models with entity type information are
a promising direction for future research on slot filling.

\section{Related Work}
\label{sec:relWork}
The slot filling shared task has been held since 2009.
There are about 20 teams participating each year.
Most systems apply a modular pipeline structure and combine
multiple approaches, such as distant supervision
and patterns \cite{overviewSF2014}.
In 2015, we were one of the first teams 
using neural networks \cite{cis2015,stanford2015,umass2015}.

In this section, different approaches for implementing the
slot filling pipeline are described, followed by a more detailed description
of two systems that are most relevant to our work: 
the top-ranked 
system from 2013 \cite{roth2013} since 
we use their distantly supervised patterns and similar features
in our support vector machines; and the 
winning system from 2015 \cite{stanford2015}
since we evaluate our system on the assessment data from 2015.
Finally, we summarize more recent developments in slot filling research.

Most slot filling systems consist of an
information retrieval-based pipeline of different 
modules. Exceptions are, for example, 
the systems by \shortciteA{indiana2014} and \citeA{stanford2015},
which rely on relational databases 
consisting of one table storing all sentences from the 
corpus and another table storing all entity mentions.
Most groups expand the query with aliases 
\shortcite<i.a.,>{roth2013,stanford2013,nyu2014,rpi2014,pris2016,umass2016}.
Our system follows this line of work and uses information retrieval
and query expansion to extract relevant documents
and cover alternate names and spelling variations.

For sentence extraction, only a subset of systems use coreference information
\shortcite<e.g.,>{nyu2014,pris2015,stanford2015,cmu2016,rpi2016,stanford2016}.
In all our experiments and analysis, coreference information
improves the final results though. \citeA{analysisRecall}
mention the long computing time of coreference resolution systems
as a major drawback. This is why we make our coreference resource publicly
available. Together with our positive results, it can help convince other
researchers to integrate this component which we consider very important,
especially for the recall of a slot filling system.
Even fewer systems apply entity linking or another form of
disambiguating different entities with the same name \shortcite<e.g.,>{talp2012,stanford2015}.
Our results with entity linking are mixed: Although it slightly
improves the final pipeline results, it leads to recall losses due to wrong links.

Especially in 2012 and 2013, many systems relied only on pattern matching
for identifying slot fillers \shortcite<i.a.,>{pris2012,talp2012,ucd2013}.
Now, more and more teams use machine learning models for slot filling relation classification,
such as naive Bayes \cite{ucd2014}, logistic regression
\cite{nyu2014,stanford2015,cmu2016,stanford2016},
conditional random fields \shortcite<e.g.,>{pris2014,cmu2015}
or support vector machines \cite<e.g.,>{roth2013,umass2015,stanford2016}.
\textsc{Mintz} and \textsc{Miml}, our baseline systems in 
Section \ref{sec:resultsBenchmark}, are used
by, i.a., \citeA{nyu2014,stanford2014} and \citeA{stanford2016}.
More recently, participants also train neural networks, 
such as bidirectional gated recurrent units \cite{pris2016},
bidirectional long short-term memory (LSTM) networks 
\cite{pris2015,stanford2015,cmu2016,rpi2016,stanford2016,umass2016}
or convolutional neural networks 
\shortcite{ibcn2014,umass2015,pris2016,stanford2016}. In 2015, we were
one of the first to show the success of neural networks,
especially convolutional neural networks, on this task
which has led to their increasing popularity.

With the exception of, e.g., \citeA{ibcn2014,nyu2014,stanford2015,cmu2016} or \citeA{stanford2016}
who use human labels or manual cleaning of noisy labels, e.g., in connection with active learning, 
the machine-learning models are trained with distant supervision. 
In this work, we approach the problem of noisy labels by an automatic self-training
procedure. A subset of the data we use for self-training includes crowdsourced annotations
but the actual cleaning process is fully automatic and uses machine learning methods.
Thus, it scales better to larger datasets than manual cleaning or manual labeling.
While some of the participants use binary models, i.a., 
\citeA{roth2013,umass2015} or \citeA{pris2016}, others train multi-class models,
i.a., \citeA{ibcn2014,stanford2015,umass2015} and \citeA{cmu2016}.
In this study, we provide a direct comparison of the performance of
binary and multi-class models. 

\citeA{cmu2016} formulate entity type constraints 
and use integer linear programming (ILP) to combine
them with relation classification. In contrast, our type-aware
models are trained end-to-end and do not rely 
on hand-crafted hard constraints. Instead, they are able to
learn correlations between entity and relation classes from data.
In the relation extraction community, the joint modeling of
entity types and relations is known to improve results
\cite<cf.,>{roth-yih2007,Yao2010joint,miwa2014}. However, only very few approaches
use neural models for joint modeling as we do in this research. 
Examples are \citeA{miwa2016} or \citeA{pawar2017} but both
of them apply their models to clean datasets which have been
manually labeled with entity types and relations.
In contrast, we conduct our experiments on distantly supervised
slot filling which provides neither clean labels for entity
nor for relation types. We show that it is possible
to use joint training of entity and relation classification
in order to reduce the problem of error propagation in
the slot filling pipeline.
The positive results of our type-aware
multi-class CNNs may motivate other researchers
in slot filling or general relation extraction to extend 
their neural models with entity type information or other features
which are known to be useful for relation classification with
traditional models.

The top-ranked system in 2013 \cite{roth2013} follows the
main trends in slot filling and applies a modular system 
based on distant supervision which is
called \textsc{RelationFactory}. Its pipeline is similar to
ours except that it uses neither entity linking 
nor coreference resolution.
Following \citeA{roth2013}, we use their distantly supervised patterns 
and add skip n-gram features to the
feature set of our support vector machines.
An important difference of our models, however, is
that we also integrate neural networks and train
not only binary models but also
multi-class models with and without entity type information.

The top-performing system in 2015 \cite{stanford2015}
uses manually labeled training data \cite{active} as well as 
a bootstrapped self-training strategy in order to avoid distant supervision.
In contrast to most other slot filling systems, they do not apply a pipeline 
system based on information
retrieval but store preprocessed versions
of all sentences and entity mentions from the source corpus
in a relational database which they access during evaluation.
As relation extractors, they apply a combination of
patterns, an open information extraction system,
logistic regression, a bidirectional long short-term memory network and special extractors
for website and alternate names slots.
In contrast to their system, we apply a traditional slot filling
pipeline based on information retrieval and train convolutional 
neural networks. In earlier work \cite{cis2015}, we also combined
convolutional and recurrent neural networks and found that adding recurrent neural
networks increased the performance only slightly. Thus, we assume that the
main reason for their better performance is the less noise
in the labels of their training data.

Last, we summarize more recent developments 
in slot filling research:
\citeA{yu2016} present a method based on trigger extraction from
dependency trees which does not require (distantly) supervised
labels and can work for any language as long as named entity recognition, 
part-of-speech tagging, dependency parsing and trigger gazetteers are available.
\citeA{lifu2017} follow our work in extracting training and
development data and in using convolutional neural networks for slot filling relation
classification. They input dependency paths into
the network and apply attention in order to account for the larger middle
contexts in slot filling relation classification. In contrast to their work, 
we extend the convolutional neural network in this paper to not only doing relation classification
but jointly learning to classify entities and relations.
Recently, \shortciteA{zhang2017} propose position-aware attention which calculates
attention weights based on the current hidden state of their LSTM, 
the output state of the LSTM and the position embeddings 
which encode the distance of the current
word to the two relation arguments. Moreover, they
publish a supervised relation extraction dataset, obtained by
crowdsourcing, for training slot filling relation classification models.
\shortciteA{chaganty2017} address the issue of evaluating new 
slot filling systems outside of the official shared task evaluations.
They build an evaluation method 
based on importance-sampling and crowdsourcing which they make publicly available.

\section{Conclusion}
\label{sec:conclusion}
In this paper, we proposed different 
type-aware convolutional neural network architectures for slot filling.
After describing our slot filling pipeline,
we focused on its relation classification component. 
Slot filling relation classification is a task with a 
particularly noisy setup due to 
distantly supervised data at training time and
error propagation through the pipeline at test time.
We were one of the first groups to show that convolutional
neural networks are successful classification models in this scenario.
We proposed three techniques of incorporating entity type information:
a pipeline-based, a joint training and a structured prediction approach.
In our experiments, we compared binary and multi-class models and showed 
that the multi-class models improved the final performance
of the slot filling pipeline. 
The model trained jointly on entity
and relation classification achieved the best micro $F_1$ scores while the 
model with structured prediction performed best in terms of macro $F_1$.
Finally, we presented several analyses to assess the impact
and errors of the different components of the pipeline, a very important
aspect which is not evaluated in the official slot filling shared task. 
Our recall analysis showed the importance
of aliases and coreference resolution. 
Our manual error analysis revealed that
the candidate extraction component (especially the named-entity-recognition
module) and the classification component are responsible for most
of the errors of the system. 
Finally, ablation studies confirmed the large positive impact of our
type-aware convolutional
neural network on the performance of the whole slot filling pipeline.

We publish our complete slot filling system, including
the source code and the presented models (\url{http://cistern.cis.lmu.de/CIS_SlotFilling}), as well as
our coreference resource (\url{http://cistern.cis.lmu.de/corefresources}) along with this paper.

\acks{
Heike Adel is a recipient of the Google  European
Doctoral Fellowship in Natural Language Processing and 
this research is supported by this fellowship.
This work is also supported by
the European Research Council
(ERC \#740516) and by
DFG (grant SCHU
2246/4-2).
}

\begin{appendix}

\section{Overview of Slots}
As described in Section \ref{sec:neuralNets-remarks}, we merge different location slots and 
inverse slots into one slot type to use our training data most effectively and avoid redundant training.
In Table \ref{tab:coveredslots}, we provide an overview of the official slot types we cover with our models.
\begin{table}[h]
\centering
\footnotesize
\begin{tabularx}{.9\textwidth}{l|>{\raggedright\arraybackslash}X}
 our label & covered original slot types \\
 \hline 
 per:age & per:age\\
 per:alternate\_names & per:alternate\_names\\
 per:cause\_of\_death & per:cause\_of\_death\\
 per:children & per:children, per:parents \\
 per:date\_of\_birth & per:date\_of\_birth\\
 per:date\_of\_death & per:date\_of\_death \\
 per:employee\_or\_member\_of & per:employee\_or\_member\_of, org:employees\_or\_members, gpe:employees\_or\_members\\
 per:location\_of\_birth & per:city\_of\_birth, per:country\_of\_birth, per:stateorprovince\_of\_birth, gpe:births\_in\_city, gpe:births\_in\_country, gpe:deaths\_in\_stateorprovince\\
 per:loc\_of\_death & per:city\_of\_death, per:country\_of\_death, per:stateorprovince\_of\_death, gpe:deaths\_in\_city, gpe:deaths\_in\_country, gpe:deaths\_in\_stateorprovince\\
 per:loc\_of\_residence & per:cities\_of\_residence, per:countries\_of\_residence, per:statesorprovinces\_of\_residence, gpe:residents\_of\_city, gpe:residents\_of\_country, gpe:residents\_of\_stateorprovince\\
 per:origin & per:origin\\
 per:schools\_attended & per:schools\_attended, org:students\\
 per:siblings & per:siblings\\
 per:spouse & per:spouse\\
 per:title & per:title\\
 org:alternate\_names & org:alternate\_names\\
 org:loc\_of\_headquarters & org:city\_of\_headquarters, org:country\_of\_headquarters, org:stateorprovince\_of\_headquarters, gpe:headquarters\_in\_city, gpe:headquarters\_in\_country, gpe:headquarters\_in\_stateorprovince\\
 org:date\_founded & org:date\_founded\\
 org:founded\_by & org:founded\_by, per:organizations\_founded, org:organizations\_founded, gpe:organizations\_founded\\
 org:members & org:members, org:member\_of, gpe:member\_of \\
 org:parents & org:parents, org:subsidiaries, gpe:subsidiaries\\
 org:top\_members\_employees & org:top\_members\_employees, per:top\_member\_employee\_of\\
\end{tabularx}
\caption{Mapping showing which of our labels for classification covers which original slot filling slot types.}
\label{tab:coveredslots}
\end{table}

For the following slots, we could not extract enough training data to train machine learning models:
per:charges, per:other\_family, per:religion, org:date\_dissolved, org:number\_of\_\-employees\_members, org:political\_religious\_affiliation, org:shareholders,  org:website, \{per, org, gpe\}:holds\_\-shares\_in.
For those slots, our slot filling pipeline uses only the pattern matcher.

 \section{Hyperparameters}
 For tuning the CNN models, we performed grid-search over the following ranges of hyperparameters:
 filter width $\in \left\{3,5\right\}$, \# conv filters $\in \left\{100,300,1000,3000\right\}$,
 hidden units for relation extraction $\in \left\{100, 300, 1000\right\}$, 
 hidden units for entity classification $\in \left\{25,100\right\}$.
 Other hyperparameters are the same for all models. For example, we use stochastic gradient descent
 with minibatches of size 10 and a learning rate of 0.1. For regularization, we add
 a L2 penalty with a weight of 1e-5.
 
 For the SVM models, we only tune $C$, the penalty of the error term.
 All the other parameters are set to the default parameters of the implementation. For example,
 the SVMs are trained with squared hinge loss and L2 regularization, the tolerance value for 
 the stopping criteria is 1e-4.
 
 \begin{table}[h]
\centering
\footnotesize
 \begin{tabular}{ll|r|r|rr||r}
 & & \multicolumn{4}{c||}{CNN} & SVM\\
     & & filter width & \# conv filters & \multicolumn{2}{c||}{\# hidden units} & \multicolumn{1}{c}{C}\\
 & & & & relation & entity & \\
  \hline
\multirow{22}{*}{\rotatebox{90}{binary}}
& per:age & 3 & 300 & 300 & - & 0.03\\
& per:alternate\_names & 5 & 300 & 300 & - & 0.03\\
& per:cause\_of\_death & 5 & 300 & 1000 & - & 1.00\\
& per:children & 3 & 300 & 100 & - & 0.01\\
& per:date\_of\_birth & 5 & 300 & 1000 & - & 0.01\\
& per:date\_of\_death & 5 & 300 & 100 & - & 0.01\\
& per:empl\_memb\_of & 3 & 300 & 100 & - & 0.01\\
& per:location\_of\_birth & 3 & 300 & 300 & - & 0.01\\ 
& per:loc\_of\_death & 3 & 300 & 300 & - & 0.01\\
& per:loc\_of\_residence & 3 & 300 & 100 & - & 0.01\\
& per:origin & 5 & 300 & 100 & - & 0.10\\
& per:schools\_att & 3 & 300 & 300 & - & 0.10\\ 
& per:siblings & 3 & 300 & 1000 & - & 0.01\\
& per:spouse & 3 & 300 & 300 & - & 0.10\\
& per:title & 3 & 300 & 100 & - & 10.00\\
& org:alternate\_names & 5 & 300 & 1000 & - & 10.00\\
& org:date\_founded & 5 & 300 & 100& - & 0.01\\
& org:founded\_by & 5 & 300 & 300& - & 0.01\\
& org:loc\_of\_headqu & 3 & 300 & 100& - & 0.01\\
& org:members & 5 & 300 & 100 & - & 3.00\\
& org:parents & 5 & 300 & 300 & - & 0.01\\
& org:top\_memb\_empl & 3 & 300 & 100 & - & 0.30\\
\hline 
\hline 
\multirow{4}{*}{\rotatebox{90}{multi-class}}
& without entity information & 3 & 3000 & 100 & - & 0.30\\
& + p & 3 & 300 & 100 & 100 & -\\
& + j & 3 & 300 & 100 &  25 & -\\
& + s & 3 & 3000 & 100 & 100 & -
\end{tabular}
\caption{Hyperparameters of CNNs and SVMs.}
\label{tab:hyperparamsCNN}
\end{table}

 Table \ref{tab:hyperparamsCNN} shows the hyperparameters tuned on the development part
 of the slot filling relation classification benchmark dataset. The configuration files which
 we use to specify the hyperparameters for the different models are included in the code
 which we publish along with this paper.

\end{appendix}

\vskip 0.2in
\bibliography{refs}
\bibliographystyle{theapa}

\end{document}